\pdfoutput=1

\documentclass[11pt]{article}

\usepackage[]{acl}

\usepackage{times}
\usepackage{latexsym}

\usepackage[T1]{fontenc}

\usepackage[utf8]{inputenc}

\usepackage{microtype}

\usepackage{graphicx}
\usepackage{amsmath}
\usepackage{amsfonts}
\usepackage{booktabs}
\usepackage{soulutf8}
\usepackage[normalem]{ulem}
\usepackage{float}
\usepackage{bm}
\usepackage{colortbl}
\usepackage{xspace}
\usepackage{arydshln}
\usepackage{multirow}
\usepackage{makecell}

\newcommand{\data}{\textsc{TempWikiBio}\xspace}

\definecolor{attryellow}{rgb}{1.0, 0.95, 0.80}

\definecolor{correctgreen}{rgb}{0.55, 0.71, 0.0}
\definecolor{incorrectred}{rgb}{0.75, 0.08, 0.08}

\definecolor{positiveblue}{rgb}{0.69, 0.88, 0.92}
\definecolor{negativeorange}{rgb}{0.98, 0.85, 0.65}

\definecolor{fullorange}{RGB}{236, 128, 60}
\definecolor{degradedgrey}{rgb}{0.85, 0.85, 0.85}

\newcommand{\hlc}[2][yellow]{{%
    \colorlet{foo}{#1}%
    \sethlcolor{foo}\hl{#2}}%
}

\title{Time-aware Prompting for Text Generation}

\author{Shuyang Cao \and Lu Wang \\
  Computer Science and Engineering \\
  University of Michigan \\
  Ann Arbor, MI \\
  \texttt{\{caoshuy, wangluxy\}@umich.edu}}

\begin{document}
\maketitle

\begin{abstract}

In this paper, we study the effects of incorporating timestamps, such as document creation dates, into generation systems.
Two types of time-aware prompts are investigated: (1) \textbf{textual prompts} that encode document timestamps in natural language sentences;
and (2) \textbf{linear prompts} that convert timestamps into continuous vectors. 
To explore extrapolation to future data points, we further introduce a new data-to-text generation dataset, \data, containing more than $4$ millions of chronologically ordered revisions of biographical articles from English Wikipedia, each paired with structured personal profiles.
Through data-to-text generation on \data, text-to-text generation on the content transfer dataset, and summarization on XSum,
we show that linear prompts on encoder and textual prompts improve the generation quality on all datasets.
Despite having less performance drop when testing on data drawn from a later time, linear prompts focus more on non-temporal information and are less sensitive to the given timestamps, according to human evaluations and sensitivity analyses.
Meanwhile, textual prompts establish the association between the given timestamps and the output dates, yielding more factual temporal information in the output.

\end{abstract}
\section{Introduction}

Temporal information, such as publication and modification dates of documents, is an inherent attribute of documents.
Both document writers and readers are aware of this information when organizing and consuming document content. For example, an event reported by a news article is likely to happen right on the publication date.
However, state-of-the-art generation models are fine-tuned from large pre-trained models without incorporating temporal information~\cite{lewis-etal-2020-bart, zhang2020pegasus}, creating a gap between document processing by humans and automatic models.
Though previous work has split datasets according to temporal information and shown deteriorated performance of large pre-trained models as knowledge becomes outdated on future data~\cite{lazaridou2021mind, jang2022towards}, it is unclear how informing models of temporal information affects generation tasks.

\begin{figure}[t]
    \centering
    \small
    \setlength{\tabcolsep}{3pt}
    \begin{tabular}{p{0.48\textwidth}}
    \toprule
        \textbf{Infobox Attributes:} name[J. Melville Broughton Jr.] \hlc[fullorange!60]{term\_start[December 31, 1948]} death\_date[March 6, 1949] ... \\
        \midrule
        \textbf{BART:} Joseph Melville Broughton Jr. (November 17, 1888 -- March 6, 1949) was the 60th Governor of the U.S. state of North Carolina from 1941 to 1945. \\
        \textbf{Linear Prompt:} Joseph Melville Broughton Jr. (November 17, 1888 -- March 6, 1949) was the 60th Governor of the U.S. state of North Carolina from 1941 to 1945 and a United States Senator \hlc[correctgreen!50]{from 1948 until his death in 1949}. \\
        \midrule
        \empty 
        \midrule
        \textbf{BBC News:} The group made a loss of \$219m (£175.1m) compared with the same time last year ... This segment posted another very strong \hlc[fullorange!60]{quarter} ... \\
        \midrule
        \textit{Original Publication Date:} 2017-02-09 \\
        \textbf{BART:} News Corp has reported a loss for \hlc[incorrectred!40]{the first three months of the year}. \\
        \textbf{Linear Prompt:} News Corp has reported a loss for \hlc[incorrectred!40]{the first three months of the year}. \\
        \textbf{Textual Prompt:} News Corp has reported a loss for \hlc[correctgreen!50]{the three months to December}. \\
        \midrule
        \textit{Date Perturbation:}
        6 months after $\mapsto$ 2017-08-09 \\
        \textbf{Textual Prompt:} News Corp has reported a loss for \hlc[correctgreen!50]{the second quarter}. \\
        \midrule
        \textit{Date Perturbation:}
        1 month after $\mapsto$ 2017-03-09 \\
        \textbf{Textual Prompt:} News Corp has reported a loss for \hlc[incorrectred!40]{the first three months of the year}. \\
        \bottomrule
    \end{tabular}
    \caption{Sample system outputs on \data for data-to-text generation and on XSum for summarization. We highlight \hlc[fullorange!60]{relevant temporal information} in the input and corresponding \hlc[correctgreen!50]{correct} (\hlc[incorrectred!40]{incorrect}) information in the model outputs. Linear prompts could encourage selecting important dates on \data, but the temporal information encoded in the linear prompts can not be captured by the model, leading to incorrect dates when resolving with the provided dates is required on XSum; while the model with textual prompts is sensitive to the provided dates and generates correct date, it lacks world knowledge (e.g., seasonal earning is only reported after the season) to handle the last case after perturbing the original publication date.
    }
    \label{fig:counterfactual_example}
\end{figure}

Therefore, this work aims to study the effects of presenting temporal information to generation models. 
Concretely, to include timestamps in model inputs, we consider prepending two types of \textbf{time-aware prompts} to the encoder or the decoder. 
First, \textbf{textual prompts} encode timestamps within natural language descriptions, as commonly used by the recent prompt engineering work~\cite{radford2019language, raffel2019exploring}. 
We further explore \textbf{linear prompts} that map timestamps to continuous vectors via linear projections.

Concretely, we fine-tune BART~\cite{lewis-etal-2020-bart} with time-aware prompts and conduct experiments on two text-to-text generation tasks: (1) content transfer~\cite{prabhumoye-etal-2019-towards} that generates the continuation of a passage using information from a given document, and (2) summarizing news articles with XSum~\cite{narayan-etal-2018-dont}. 
To study time-aware prompts' capability of extrapolating to future dates, 
we introduce \textbf{\data}, a data-to-text dataset containing timestamped revisions of biographical articles from English Wikipedia, each paired with an infobox as input. The revisions record changes of personal profiles from 2004 to 2021.\footnote{Our data and code are available at \url{https://shuyangcao.github.io/projects/temporal_prompt_generation}.}
For all experimented datasets, dated events are critical.
We first evaluate model outputs with automatic metrics to examine the effects of temporal information on output informativeness.
Human judges are then asked to additionally rate the factuality of model outputs and determine if the improvement or degradation is due to date changes in the outputs.
Finally, we perform a sensitivity analysis by making perturbations to the original dates (e.g., setting the dates to one year before) and providing models with the perturbed dates, and then inspect the changes of outputs. We find that:

\begin{itemize}
    \item Time-aware prompts improve the model performance over the no-prompt baseline in 87.5\% of comparisons on different metrics and datasets. Linear prompts work better on the data-to-text dataset, while textual prompts work better on the text-to-text datasets, partly due to modal compatibility. 
    \item The improvement in output informativeness and factuality by linear prompts is less frequently related to modifying temporal information in the outputs than textual prompts, according to human judges. Moreover, models with linear prompts are less sensitive to the given temporal information, suggesting that linear prompts assist the processing of non-temporal content. 
    \item Textual prompts associate the provided dates with the dates to be generated in the outputs, producing more factual time-related information. However, models with textual prompts could generate incorrect dates when complicated world knowledge is required to perform reasoning, as shown in the last example in Figure~\ref{fig:counterfactual_example}. 
\end{itemize}

\section{Related Work}
\label{sec:related_work}

\paragraph{Temporal Generalization in NLP.}

Early work on temporal generalization focuses on detecting the shifts of n-gram frequencies over time~\cite{doi:10.1126/science.1199644} and detecting word meaning changes~\cite{10.1145/2064448.2064475,kulkarni2015statistically}. 
Besides linguistic shifts, model degradation on downstream tasks has been reported when tested on samples at a different time from the training data~\cite{huang-paul-2018-examining, lukes-sogaard-2018-sentiment, lazaridou2021mind, DBLP:journals/corr/abs-2111-12790}. 
In this work, we study the temporal generalization of our time-aware prompts, since they are constructed with temporal information.

\paragraph{Prompt Engineering.} 

Prompts have been a common tool for controllable generation~\cite{fan-etal-2018-controllable,radford2019language, keskar2019ctrl, raffel2019exploring}. 
Instructions are also constructed as prompts to allow large models to perform new tasks that are unseen in training~\cite{brown2020language, sanh2022multitask}. 
More recently, prompts, either hand-crafted~\cite{schick-schutze-2021-just, gao-etal-2021-making} or learned~\cite{li-liang-2021-prefix}, are found to benefit model learning and improve few-shot performance on downstream tasks. 
Our textual time-aware prompts extend the year-level prompts in \citet{dhingra2021timeaware} with months and days to incorporate fine-grained temporal information, and we further explore representing timestamps with linear prompts which have been mainly used for length-controlled generation~\cite{kikuchi-etal-2016-controlling}.
\section{Time-aware Prompts}

We study two types of prompts that are prepended to the encoder/decoder of a seq2seq model, to inform the model of temporal information.

\paragraph{Textual Prompt.}
Given a document's timestamp, we first convert it to ``day month year'' with the day and the year in digits and the month in its textual form (e.g., ``\textit{18 January 2015}''), a format commonly used by mainstream media such as BBC news. 
We test three textual prompts and use the one that results in the highest ROUGE score on the development set of XSum, i.e., ``\textit{Today is [timestamp].}'' (``\textit{Today is 18 January 2015.}'') 
Other textual prompts are detailed in Appendix~\ref{appendix:prompt}. 
Compared to only inserting the year information~\cite{dhingra2021timeaware}, textual prompts in our paper provide more fine-grained temporal information.

\begin{figure}[t]
    \centering
    \includegraphics[width=0.48\textwidth]{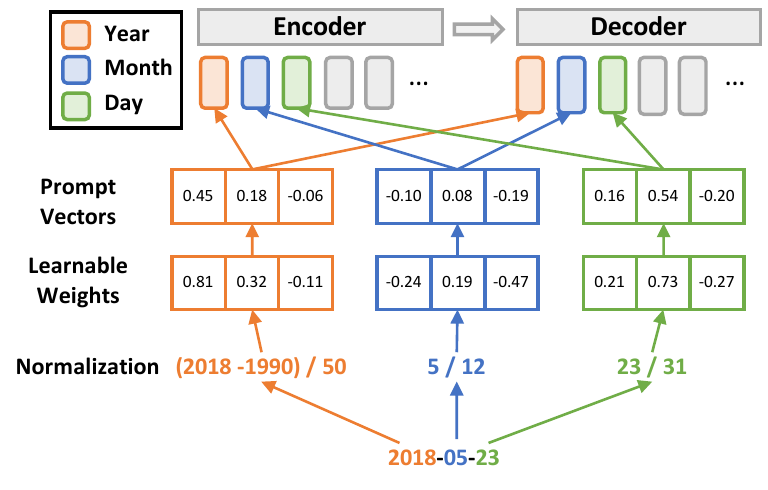}
    \caption{A linear prompt treats year/month/day as separate scalars and projects them into continuous prompt vectors to be used on encoder or decoder. The vector's scale reflects their temporal orderings.
    Note that the dimension of prompt vectors is the same as the embedding dimension in the actual model.}
    \label{fig:linear_time_prompt}
\end{figure}

\smallskip
\noindent\textbf{Linear Prompts} 
treat the concept of time as an axis, with each timestamp being mapped to a point on it.
Concretely, we use the year, month, and day as scalars and transform them into prompt vectors through linear projections, as illustrated in Figure~\ref{fig:linear_time_prompt}. The process of linear projections can also be viewed as changing the scales of vectors for the year, month, and day.
While prior work has controlled the output lengths by changing the scales of memory cells in an LSTM~\cite{kikuchi-etal-2016-controlling}, representing temporal information with scales of vectors has yet been studied.

\section{\data Data Collection}

\begin{figure}
    \centering
    \includegraphics[width=0.45\textwidth]{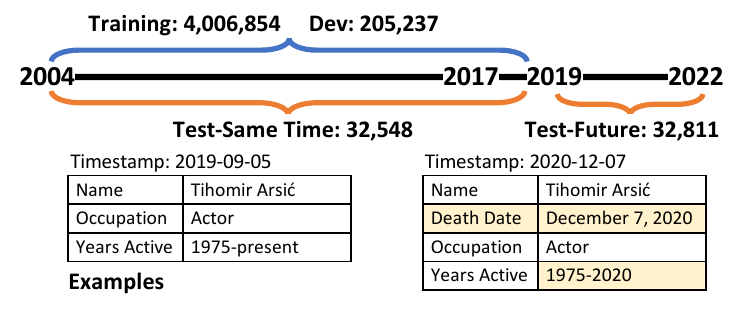}
    \caption{Numbers of samples and the corresponding time periods of revisions in different splits of \data. Changed attributes of the sample revisions are shaded in \hlc[attryellow]{yellow}. 
    There is \textbf{no} overlapping subject between training/development sets and the two test sets. 
    }
    \label{fig:wiki_rev_diff}
\end{figure}

To study how well time-aware prompts can extrapolate to future data, we collect \data, which has $4{,}277{,}450$ revisions of infobox-paragraph pairs from 2004 to 2021 for $695{,}929$ Wikipedia biography articles, extending \textsc{WikiBio}~\cite{lebret-etal-2016-neural}, which only includes the latest revision per article by 2015. 
Importantly, the profile (e.g., titles, awards, etc) of a person shown in the infobox changes over time (Figure~\ref{fig:wiki_rev_diff}). 

Concretely, for each biography, we pick its latest revision every X days since the first revision, where X is sampled uniformly from $[270, 450]$, to diversify the timestamps included in the data. 
We then extract the infobox and the lead paragraph per revision. 
As illustrated in Figure~\ref{fig:wiki_rev_diff}, two test sets are created:
\texttt{test-same time} contains articles that are published at the same time as the training set, while \texttt{test-future} consists of samples that are created (or revised) after training and development sets.
We further ensure that the subjects of biographies in both test sets are not in training or development sets. 
On average, each revision has $15.3$ attributes in the infobox and $43.2$ words in the first paragraph. 
Details of data collection are included in Appendix~\ref{appendix:dataset}.

\section{Experiments and Results}

For data-to-text generation on \data, we linearize the infobox and use it as the input to BART. 
Common data-to-text metrics~\cite{gehrmann-etal-2021-gem} are used, including BLEU-4~\cite{papineni-etal-2002-bleu}, METEOR~\cite{lavie-agarwal-2007-meteor}, TER~\cite{snover-etal-2006-study}, and BERTScore~\cite{Zhang2020BERTScore}.
For text-to-text summarization on XSum~\cite{narayan-etal-2018-dont} that consists of news articles and their corresponding summaries from BBC News, we report ROUGE scores~\cite{lin-2004-rouge} and QuestEval~\cite{scialom-etal-2021-questeval}, a QA-based faithfulness evaluation metric that checks if questions created from the summary can be addressed by reading the document with a QA model, and vice versa. 
%
The content transfer dataset~\cite{prabhumoye-etal-2019-towards} considers sentences containing citations of news sources in Wikipedia articles as the target for generation. 
Many target sentences incorporate important dates from the cited articles, thus making it suitable to test our time-aware prompt design. 
To generate each target sentence, the context passage, which contains three sentences preceding the target sentence, and the cited news article are provided as input.
ROUGE-L, METEOR, and BERTScore are computed for evaluation on the content transfer dataset.

\begin{table}[t]
    \centering
    \small
    \setlength{\tabcolsep}{4pt}
    \begin{tabular}{clllll}
    \toprule
        & \textbf{Prompt} & \textbf{B-4} ($\uparrow$) & \textbf{MTR} ($\uparrow$) & \textbf{TER} ($\downarrow$) & \textbf{BS} ($\uparrow$) \\
        \midrule
        \multirowcell{5}{\textbf{\textit{Test}}\\\textbf{\textit{Future}}} & - & 30.48 & 49.88 & 69.66 & 55.34 \\
        \cdashline{2-6}
        & \textsc{Enc:T} & \hlc[positiveblue]{30.76}$^\ast$ & \hlc[positiveblue]{50.10}$^\ast$ & \hlc[positiveblue]{69.33} & \hlc[positiveblue]{55.38} \\
        & \textsc{Enc:L} & \hlc[positiveblue]{\textbf{31.22}}$^\ast$ & \hlc[positiveblue]{\textit{50.32}}$^\ast$ & \hlc[positiveblue]{\textbf{68.66}}$^\ast$ & \hlc[positiveblue]{\textbf{55.79}}$^\ast$ \\
        & \textsc{Dec:T} & \hlc[positiveblue]{\textit{31.05}}$^\ast$ & \hlc[positiveblue]{\textbf{50.46}}$^\ast$ & 69.98 & \hlc[positiveblue]{55.35} \\
        & \textsc{Dec:L} & \hlc[positiveblue]{30.69}$^\ast$ & \hlc[positiveblue]{49.94} & \hlc[positiveblue]{\textit{69.03}}$^\ast$ & \hlc[positiveblue]{\textit{\textit{55.52}}}$^\ast$ \\
        \midrule
        \multirowcell{5}{\textbf{\textit{Test}}\\\textbf{\textit{Same}}\\\textbf{\textit{Time}}} & - & 30.81 & 50.15 & 70.26 & 55.51 \\
        \cdashline{2-6}
        & \textsc{Enc:T} & \hlc[positiveblue]{31.27}$^\ast$ & \hlc[positiveblue]{50.55}$^\ast$ & \hlc[positiveblue]{\textit{69.78}}$^\ast$ & \hlc[positiveblue]{\textit{55.82}}$^\ast$ \\
        & \textsc{Enc:L} & \hlc[positiveblue]{\textbf{31.50}}$^\ast$ & \hlc[positiveblue]{\textit{50.56}}$^\ast$ & \hlc[positiveblue]{\textbf{69.33}}$^\ast$ & \hlc[positiveblue]{\textbf{56.00}}$^\ast$ \\
        & \textsc{Dec:T} & \hlc[positiveblue]{\textit{31.43}}$^\ast$ & \textbf{\hlc[positiveblue]{50.73}}$^\ast$ & 70.51 & \hlc[positiveblue]{55.62} \\
        & \textsc{Dec:L} & \hlc[positiveblue]{30.91} & 50.11 & \hlc[positiveblue]{69.86} & \hlc[positiveblue]{55.59} \\
        \bottomrule
    \end{tabular}
    \caption{
    Results on \data. Textual (T) and linear (L) prompts are used on encoder (\textsc{Enc}) or decoder (\textsc{Dec}). 
    B: BLEU; MTR: METEOR; BS: BERTScore. The best result per metric is in \textbf{boldface} and the second best is in \textit{italics}.
    Improvement over the no-prompt baseline is \hlc[positiveblue]{shaded}.
    $^\ast$: significantly better than the baseline with approximate randomization test ($p < 0.001$). 
    }
    \label{tab:wiki_auto_eval}
\end{table}

\begin{table}[t]
    \centering
    \small
    \setlength{\tabcolsep}{3.3pt}
    \begin{tabular}{cllllll}
    \toprule
        & \multicolumn{3}{c}{\textbf{Content Transfer}} & \multicolumn{3}{c}{\textbf{XSum}} \\
        \cmidrule(lr){2-4}\cmidrule(lr){5-7}
        \textbf{Prompt} & \textbf{R-L} & \textbf{MTR} & \textbf{BS} & \textbf{R-1} & \textbf{R-2} & \textbf{QEval} \\
        \midrule
        - & 27.52 & 27.55 & 29.86 & 45.23 & 22.11 & 47.73 \\
        \hdashline
        \textsc{Enc:T} & \hlc[positiveblue]{\textit{28.15}}$^\ast$ & \hlc[positiveblue]{\textit{28.40}}$^\ast$ & \hlc[positiveblue]{\textit{30.59}}$^\ast$ & \hlc[positiveblue]{\textbf{45.63}}$^\ast$ & \hlc[positiveblue]{\textit{22.38}}$^\ast$ & \hlc[positiveblue]{47.76} \\
        \textsc{Enc:L} & \hlc[positiveblue]{27.82}$^\ast$ & \hlc[positiveblue]{27.99}$^\ast$ & \hlc[positiveblue]{30.33}$^\ast$ & \hlc[positiveblue]{45.32} & \hlc[positiveblue]{22.22} & \hlc[positiveblue]{47.76} \\
        \textsc{Dec:T} & \hlc[positiveblue]{\textbf{28.41}}$^\ast$ & \hlc[positiveblue]{\textbf{28.84}}$^\ast$ & \hlc[positiveblue]{\textbf{30.90}}$^\ast$ & \hlc[positiveblue]{\textit{45.59}}$^\ast$ & \hlc[positiveblue]{\textbf{22.45}}$^\ast$ & 47.70 \\
        \textsc{Dec:L} & \hlc[positiveblue]{27.62} & \hlc[positiveblue]{27.68} & \hlc[positiveblue]{30.13}$^\ast$ & 44.94 & 21.91 & 47.51 \\
        \bottomrule
    \end{tabular}
    \caption{Results on the content transfer and XSum datasets. R: ROUGE; QEval: QuestEval.
    }
    \label{tab:cf_auto_eval}
\end{table}

\paragraph{Automatic Evaluation.}
Overall, models with time-aware prompts obtain better performance than the no-prompt baseline. Models with time-aware prompts win 49 of all 56 comparisons against the baseline on different metrics and datasets, indicating that adding time-aware prompts encourages the models to generate more informative outputs.

Linear prompts tend to work better on the encoder when the input is structured data, achieving the best overall performance on \data (Table~\ref{tab:wiki_auto_eval}). Linear prompts also show less performance degradation than textual prompts when testing future samples.
However, the better extrapolation performance by the model with linear prompts might be due to its lower sensitivity to the provided dates, as later revealed in our analysis.

When the input and output are both in natural language, textual prompts are more suitable, as evidenced by the better performance than linear prompts on all metrics on both the content transfer dataset and XSum (Table~\ref{tab:cf_auto_eval}).
We think that on text-to-text generation tasks, textual prompts benefit from modal compatibility and have an advantage of connecting the salient content with the timestamps.

\paragraph{Human Evaluation.}
We hire three fluent English speakers to evaluate $80$ sets of paragraphs generated for \data samples at a future time  and $80$ sets of sentences generated for content transfer samples by two models with time-aware prompts. 
The judges compare the output by each model against the output by the no-prompt baseline on two aspects: \textbf{informativeness} --- whether the model output covers salient information in the input; and \textbf{factuality} --- whether the model output is factually correct.
For each set, the judges only know which output is generated by the baseline, and outputs by other models are randomly sorted. 
Besides the three-way label (win/tie/lose), we ask the judges to determine if the difference in each aspect between each pair of comparisons is date-related. 
\begin{figure}
    \centering
    \includegraphics[width=0.48\textwidth]{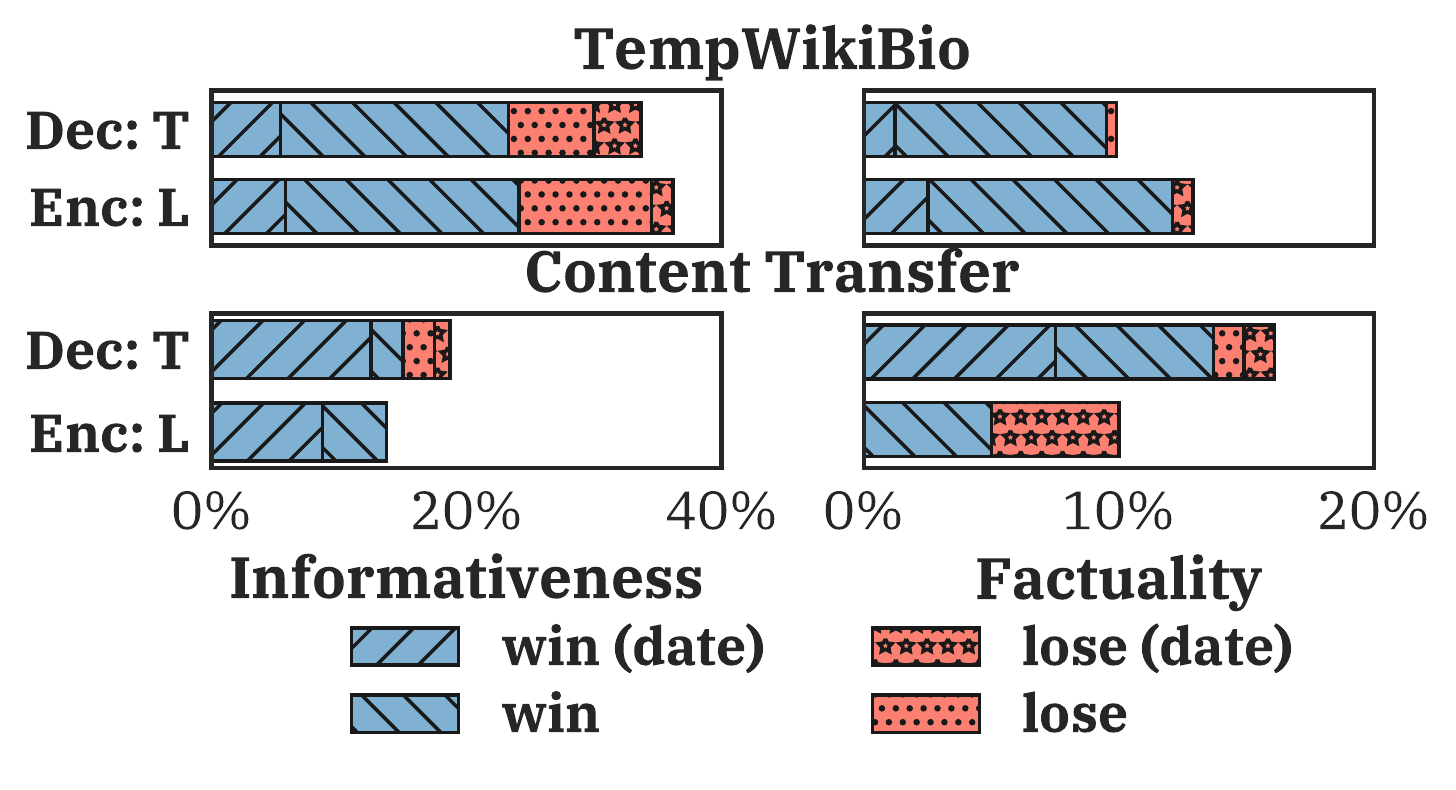}
    \caption{Percentages of samples that win or lose over the no-prompt baseline, on Test-Future of \data and the content transfer dataset.
    While both time-aware prompts improve informativeness and factuality on \data, textual prompts are more often rated as having date-related improvements in informativeness and factuality on the content transfer dataset.
    Krippendorff's $\alpha$: 0.85 (informativeness); 0.64 (factuality).}
    \label{fig:human_eval_result}
\end{figure}

As shown in Figure~\ref{fig:human_eval_result}, only a small portion of improvements by linear prompts is date-related, especially on the content transfer dataset where none of the outputs by the model with linear prompts is rated as having better factuality due to the inclusions or changes of dates, suggesting that linear prompts might help process other content.
By contrast, the model with textual prompts focuses more on the temporal information and brings more factual dates into the outputs on the content transfer dataset.

\begin{figure}
    \centering
    \includegraphics[width=0.48\textwidth]{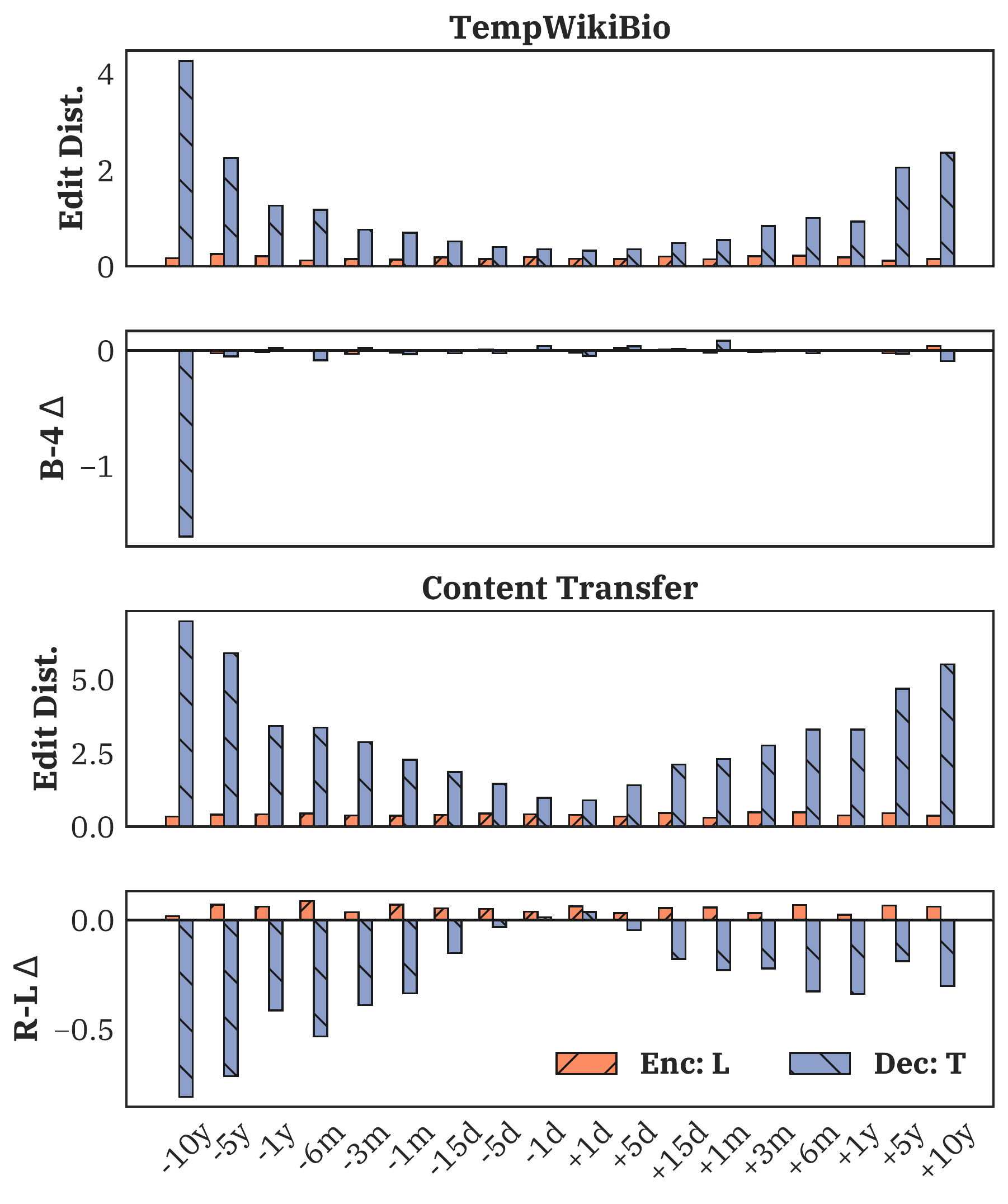}
    \caption{Edit distances and differences of BLEU-4/ROUGE-L between outputs with perturbed dates and original dates on \data and content transfer. Linear prompts are not sensitive to the given dates.
    Results on XSum are in Appendix~\ref{appendix:additional_results}.
    }
    \label{fig:intervention_wiki_cf_result}
\end{figure}

\paragraph{Analysis via Date Perturbation.}
We probe into date sensitivity, to understand the mechanisms behind the two types of prompts. Specifically, the original timestamps of $2000$ samples randomly selected from the test set of each dataset are perturbed and provided to the models.
As indicated by the greater edit distances between the outputs produced with the perturbed dates and original dates (Figure~\ref{fig:intervention_wiki_cf_result}), models with textual prompts are more sensitive to the given dates than models with linear prompts.
Human inspection of the outputs by the model with linear prompts and perturbed dates also finds that their changes from the original outputs are not related to the temporal information, echoing that the improvements in informativeness and factuality are less date-related according to human judges.
For the model with textual prompts, we observe a need for learning complicated world knowledge to generate correct dates more frequently when provided perturbed dates, as shown in Figure~\ref{fig:counterfactual_example} and Figure~\ref{fig:example_xsum} in Appendix~\ref{appendix:example_output}.

In addition, the greater differences of ROUGE-L scores suggest a more significant dependency on the temporal information by the content transfer dataset, where the publication dates of documents are often required to generate the outputs (Figure~\ref{fig:example_cf} in Appendix~\ref{appendix:example_output}), calling for the inclusion of metadata during data collection.

\section{Conclusion}

We study two types of time-aware prompts for injecting document timestamps into generation models.
Experiments on \data, our newly collected data-to-text generation dataset, and two text-to-text generation tasks show that linear prompts mostly enhance the processing of content other than dates for more informative and factual outputs.
Textual prompts build the association between the given temporal information and the generated temporal information, producing outputs with more factual dates.

\section*{Acknowledgements}

This work is supported in part by National Science Foundation through grant IIS-2046016, and Oracle Cloud credits and related resources provided by the Oracle for Research program. 
We thank the anonymous reviewers for their valuable suggestions.

\section*{Ethical Consideration}

Our work assumes the timestamps of documents can be accurately obtained and the models are always provided with the accurate creation dates. 
However, this might not be the case for some documents, especially the ones that are first published in a paper format and later digitized into electronic versions. 
Informing generation models of inaccurate timestamps could lead to incorrect content generation and other unpredictable behaviors, where fabricated facts might be picked up by end users, potentially causing harm to the public.

\section*{Limitation}

Though we show that textual time-aware prompts help models generate more factually consistent outputs, we find that models with temporal prompts could generate incorrect temporal information due to the lack of world knowledge (Figure~\ref{fig:example_xsum} of Appendix~\ref{appendix:example_output}).
In this work, we do not further study methods that can incorporate extra world knowledge to address this issue.

During model evaluation, we investigate the effects of time-aware prompts on the generated temporal information via human evaluation, which includes 160 outputs by each model (320 in total).
We believe automatic metrics that verify the correctness of temporal information in the outputs can better validate the improvements by our models.
However, such automatic metrics do not exist. A potential design of temporal information evaluation metrics is to combine event and temporal expression extraction systems.
We made several attempts at this design, but the performance of the event and temporal expression extraction systems we tested needs further improvement.

To obtain the timestamp of each sample, we rely on the automatic web archive (Appendix~\ref{appendix:dataset}). However, this approach for timestamp retrieval only applies to datasets that are based on web sources (e.g., news articles and blog posts). In addition, less popular web sources are less likely to be archived by automatic web archive service, which makes retrieving their timestamps more complicated and prevents the adoption of our methods.

\bibliography{custom}

\begin{thebibliography}{36}
\expandafter\ifx\csname natexlab\endcsname\relax\def\natexlab#1{#1}\fi

\bibitem[{Agarwal and Nenkova(2021)}]{DBLP:journals/corr/abs-2111-12790}
Oshin Agarwal and Ani Nenkova. 2021.
\newblock \href {http://arxiv.org/abs/2111.12790} {Temporal effects on
  pre-trained models for language processing tasks}.
\newblock \emph{CoRR}, abs/2111.12790.

\bibitem[{Bird et~al.(2009)Bird, Klein, and Loper}]{bird2009natural}
Steven Bird, Ewan Klein, and Edward Loper. 2009.
\newblock \emph{Natural language processing with Python: analyzing text with
  the natural language toolkit}.
\newblock " O'Reilly Media, Inc.".

\bibitem[{Brown et~al.(2020)Brown, Mann, Ryder, Subbiah, Kaplan, Dhariwal,
  Neelakantan, Shyam, Sastry, Askell et~al.}]{brown2020language}
Tom Brown, Benjamin Mann, Nick Ryder, Melanie Subbiah, Jared~D Kaplan, Prafulla
  Dhariwal, Arvind Neelakantan, Pranav Shyam, Girish Sastry, Amanda Askell,
  et~al. 2020.
\newblock Language models are few-shot learners.
\newblock \emph{Advances in neural information processing systems},
  33:1877--1901.

\bibitem[{Dhingra et~al.(2021)Dhingra, Cole, Eisenschlos, Gillick, Eisenstein,
  and Cohen}]{dhingra2021timeaware}
Bhuwan Dhingra, Jeremy~R. Cole, Julian~Martin Eisenschlos, Daniel Gillick,
  Jacob Eisenstein, and William~W. Cohen. 2021.
\newblock \href {http://arxiv.org/abs/2106.15110} {Time-aware language models
  as temporal knowledge bases}.

\bibitem[{Dhingra et~al.(2019)Dhingra, Faruqui, Parikh, Chang, Das, and
  Cohen}]{dhingra-etal-2019-handling}
Bhuwan Dhingra, Manaal Faruqui, Ankur Parikh, Ming-Wei Chang, Dipanjan Das, and
  William Cohen. 2019.
\newblock \href {https://doi.org/10.18653/v1/P19-1483} {Handling divergent
  reference texts when evaluating table-to-text generation}.
\newblock In \emph{Proceedings of the 57th Annual Meeting of the Association
  for Computational Linguistics}, pages 4884--4895, Florence, Italy.
  Association for Computational Linguistics.

\bibitem[{Fan et~al.(2018)Fan, Grangier, and Auli}]{fan-etal-2018-controllable}
Angela Fan, David Grangier, and Michael Auli. 2018.
\newblock \href {https://doi.org/10.18653/v1/W18-2706} {Controllable
  abstractive summarization}.
\newblock In \emph{Proceedings of the 2nd Workshop on Neural Machine
  Translation and Generation}, pages 45--54, Melbourne, Australia. Association
  for Computational Linguistics.

\bibitem[{Gao et~al.(2021)Gao, Fisch, and Chen}]{gao-etal-2021-making}
Tianyu Gao, Adam Fisch, and Danqi Chen. 2021.
\newblock \href {https://doi.org/10.18653/v1/2021.acl-long.295} {Making
  pre-trained language models better few-shot learners}.
\newblock In \emph{Proceedings of the 59th Annual Meeting of the Association
  for Computational Linguistics and the 11th International Joint Conference on
  Natural Language Processing (Volume 1: Long Papers)}, pages 3816--3830,
  Online. Association for Computational Linguistics.

\bibitem[{Gehrmann et~al.(2021)Gehrmann, Adewumi, Aggarwal, Ammanamanchi,
  Aremu, Bosselut, Chandu, Clinciu, Das, Dhole, Du, Durmus, Du{\v{s}}ek,
  Emezue, Gangal, Garbacea, Hashimoto, Hou, Jernite, Jhamtani, Ji, Jolly, Kale,
  Kumar, Ladhak, Madaan, Maddela, Mahajan, Mahamood, Majumder, Martins,
  McMillan-Major, Mille, van Miltenburg, Nadeem, Narayan, Nikolaev,
  Niyongabo~Rubungo, Osei, Parikh, Perez-Beltrachini, Rao, Raunak, Rodriguez,
  Santhanam, Sedoc, Sellam, Shaikh, Shimorina, Sobrevilla~Cabezudo, Strobelt,
  Subramani, Xu, Yang, Yerukola, and Zhou}]{gehrmann-etal-2021-gem}
Sebastian Gehrmann, Tosin Adewumi, Karmanya Aggarwal, Pawan~Sasanka
  Ammanamanchi, Anuoluwapo Aremu, Antoine Bosselut, Khyathi~Raghavi Chandu,
  Miruna-Adriana Clinciu, Dipanjan Das, Kaustubh Dhole, Wanyu Du, Esin Durmus,
  Ond{\v{r}}ej Du{\v{s}}ek, Chris~Chinenye Emezue, Varun Gangal, Cristina
  Garbacea, Tatsunori Hashimoto, Yufang Hou, Yacine Jernite, Harsh Jhamtani,
  Yangfeng Ji, Shailza Jolly, Mihir Kale, Dhruv Kumar, Faisal Ladhak, Aman
  Madaan, Mounica Maddela, Khyati Mahajan, Saad Mahamood, Bodhisattwa~Prasad
  Majumder, Pedro~Henrique Martins, Angelina McMillan-Major, Simon Mille, Emiel
  van Miltenburg, Moin Nadeem, Shashi Narayan, Vitaly Nikolaev, Andre
  Niyongabo~Rubungo, Salomey Osei, Ankur Parikh, Laura Perez-Beltrachini,
  Niranjan~Ramesh Rao, Vikas Raunak, Juan~Diego Rodriguez, Sashank Santhanam,
  Jo{\~a}o Sedoc, Thibault Sellam, Samira Shaikh, Anastasia Shimorina,
  Marco~Antonio Sobrevilla~Cabezudo, Hendrik Strobelt, Nishant Subramani, Wei
  Xu, Diyi Yang, Akhila Yerukola, and Jiawei Zhou. 2021.
\newblock \href {https://doi.org/10.18653/v1/2021.gem-1.10} {The {GEM}
  benchmark: Natural language generation, its evaluation and metrics}.
\newblock In \emph{Proceedings of the 1st Workshop on Natural Language
  Generation, Evaluation, and Metrics (GEM 2021)}, pages 96--120, Online.
  Association for Computational Linguistics.

\bibitem[{Huang and Paul(2018)}]{huang-paul-2018-examining}
Xiaolei Huang and Michael~J. Paul. 2018.
\newblock \href {https://doi.org/10.18653/v1/P18-2110} {Examining temporality
  in document classification}.
\newblock In \emph{Proceedings of the 56th Annual Meeting of the Association
  for Computational Linguistics (Volume 2: Short Papers)}, pages 694--699,
  Melbourne, Australia. Association for Computational Linguistics.

\bibitem[{Jang et~al.(2022)Jang, Ye, Yang, Shin, Han, KIM, Choi, and
  Seo}]{jang2022towards}
Joel Jang, Seonghyeon Ye, Sohee Yang, Joongbo Shin, Janghoon Han, Gyeonghun
  KIM, Stanley~Jungkyu Choi, and Minjoon Seo. 2022.
\newblock \href {https://openreview.net/forum?id=vfsRB5MImo9} {Towards
  continual knowledge learning of language models}.
\newblock In \emph{International Conference on Learning Representations}.

\bibitem[{Keskar et~al.(2019)Keskar, McCann, Varshney, Xiong, and
  Socher}]{keskar2019ctrl}
Nitish~Shirish Keskar, Bryan McCann, Lav~R. Varshney, Caiming Xiong, and
  Richard Socher. 2019.
\newblock \href {http://arxiv.org/abs/1909.05858} {Ctrl: A conditional
  transformer language model for controllable generation}.

\bibitem[{Kikuchi et~al.(2016)Kikuchi, Neubig, Sasano, Takamura, and
  Okumura}]{kikuchi-etal-2016-controlling}
Yuta Kikuchi, Graham Neubig, Ryohei Sasano, Hiroya Takamura, and Manabu
  Okumura. 2016.
\newblock \href {https://doi.org/10.18653/v1/D16-1140} {Controlling output
  length in neural encoder-decoders}.
\newblock In \emph{Proceedings of the 2016 Conference on Empirical Methods in
  Natural Language Processing}, pages 1328--1338, Austin, Texas. Association
  for Computational Linguistics.

\bibitem[{Kulkarni et~al.(2015)Kulkarni, Al-Rfou, Perozzi, and
  Skiena}]{kulkarni2015statistically}
Vivek Kulkarni, Rami Al-Rfou, Bryan Perozzi, and Steven Skiena. 2015.
\newblock Statistically significant detection of linguistic change.
\newblock In \emph{Proceedings of the 24th international conference on world
  wide web}, pages 625--635.

\bibitem[{Lavie and Agarwal(2007)}]{lavie-agarwal-2007-meteor}
Alon Lavie and Abhaya Agarwal. 2007.
\newblock \href {https://aclanthology.org/W07-0734} {{METEOR}: An automatic
  metric for {MT} evaluation with high levels of correlation with human
  judgments}.
\newblock In \emph{Proceedings of the Second Workshop on Statistical Machine
  Translation}, pages 228--231, Prague, Czech Republic. Association for
  Computational Linguistics.

\bibitem[{Lazaridou et~al.(2021)Lazaridou, Kuncoro, Gribovskaya, Agrawal,
  Liska, Terzi, Gimenez, de~Masson~d'Autume, Kocisky, Ruder
  et~al.}]{lazaridou2021mind}
Angeliki Lazaridou, Adhi Kuncoro, Elena Gribovskaya, Devang Agrawal, Adam
  Liska, Tayfun Terzi, Mai Gimenez, Cyprien de~Masson~d'Autume, Tomas Kocisky,
  Sebastian Ruder, et~al. 2021.
\newblock Mind the gap: Assessing temporal generalization in neural language
  models.
\newblock \emph{Advances in Neural Information Processing Systems}, 34.

\bibitem[{Lebret et~al.(2016)Lebret, Grangier, and
  Auli}]{lebret-etal-2016-neural}
R{'e}mi Lebret, David Grangier, and Michael Auli. 2016.
\newblock \href {https://doi.org/10.18653/v1/D16-1128} {Neural text generation
  from structured data with application to the biography domain}.
\newblock In \emph{Proceedings of the 2016 Conference on Empirical Methods in
  Natural Language Processing}, pages 1203--1213, Austin, Texas. Association
  for Computational Linguistics.

\bibitem[{Lewis et~al.(2020)Lewis, Liu, Goyal, Ghazvininejad, Mohamed, Levy,
  Stoyanov, and Zettlemoyer}]{lewis-etal-2020-bart}
Mike Lewis, Yinhan Liu, Naman Goyal, Marjan Ghazvininejad, Abdelrahman Mohamed,
  Omer Levy, Veselin Stoyanov, and Luke Zettlemoyer. 2020.
\newblock \href {https://doi.org/10.18653/v1/2020.acl-main.703} {{BART}:
  Denoising sequence-to-sequence pre-training for natural language generation,
  translation, and comprehension}.
\newblock In \emph{Proceedings of the 58th Annual Meeting of the Association
  for Computational Linguistics}, pages 7871--7880, Online. Association for
  Computational Linguistics.

\bibitem[{Li and Liang(2021)}]{li-liang-2021-prefix}
Xiang~Lisa Li and Percy Liang. 2021.
\newblock \href {https://doi.org/10.18653/v1/2021.acl-long.353} {Prefix-tuning:
  Optimizing continuous prompts for generation}.
\newblock In \emph{Proceedings of the 59th Annual Meeting of the Association
  for Computational Linguistics and the 11th International Joint Conference on
  Natural Language Processing (Volume 1: Long Papers)}, pages 4582--4597,
  Online. Association for Computational Linguistics.

\bibitem[{Lin(2004)}]{lin-2004-rouge}
Chin-Yew Lin. 2004.
\newblock \href {https://aclanthology.org/W04-1013} {{ROUGE}: A package for
  automatic evaluation of summaries}.
\newblock In \emph{Text Summarization Branches Out}, pages 74--81, Barcelona,
  Spain. Association for Computational Linguistics.

\bibitem[{Lukes and S{\o}gaard(2018)}]{lukes-sogaard-2018-sentiment}
Jan Lukes and Anders S{\o}gaard. 2018.
\newblock \href {https://doi.org/10.18653/v1/W18-6210} {Sentiment analysis
  under temporal shift}.
\newblock In \emph{Proceedings of the 9th Workshop on Computational Approaches
  to Subjectivity, Sentiment and Social Media Analysis}, pages 65--71,
  Brussels, Belgium. Association for Computational Linguistics.

\bibitem[{Michel et~al.(2011)Michel, Shen, Aiden, Veres, Gray, null null,
  Pickett, Hoiberg, Clancy, Norvig, Orwant, Pinker, Nowak, and
  Aiden}]{doi:10.1126/science.1199644}
Jean-Baptiste Michel, Yuan~Kui Shen, Aviva~Presser Aiden, Adrian Veres,
  Matthew~K. Gray, null null, Joseph~P. Pickett, Dale Hoiberg, Dan Clancy,
  Peter Norvig, Jon Orwant, Steven Pinker, Martin~A. Nowak, and Erez~Lieberman
  Aiden. 2011.
\newblock \href {https://doi.org/10.1126/science.1199644} {Quantitative
  analysis of culture using millions of digitized books}.
\newblock \emph{Science}, 331(6014):176--182.

\bibitem[{Narayan et~al.(2018)Narayan, Cohen, and
  Lapata}]{narayan-etal-2018-dont}
Shashi Narayan, Shay~B. Cohen, and Mirella Lapata. 2018.
\newblock \href {https://doi.org/10.18653/v1/D18-1206} {Don{'}t give me the
  details, just the summary! topic-aware convolutional neural networks for
  extreme summarization}.
\newblock In \emph{Proceedings of the 2018 Conference on Empirical Methods in
  Natural Language Processing}, pages 1797--1807, Brussels, Belgium.
  Association for Computational Linguistics.

\bibitem[{Ott et~al.(2019)Ott, Edunov, Baevski, Fan, Gross, Ng, Grangier, and
  Auli}]{ott2019fairseq}
Myle Ott, Sergey Edunov, Alexei Baevski, Angela Fan, Sam Gross, Nathan Ng,
  David Grangier, and Michael Auli. 2019.
\newblock fairseq: A fast, extensible toolkit for sequence modeling.
\newblock In \emph{Proceedings of NAACL-HLT 2019: Demonstrations}.

\bibitem[{Papineni et~al.(2002)Papineni, Roukos, Ward, and
  Zhu}]{papineni-etal-2002-bleu}
Kishore Papineni, Salim Roukos, Todd Ward, and Wei-Jing Zhu. 2002.
\newblock \href {https://doi.org/10.3115/1073083.1073135} {{B}leu: a method for
  automatic evaluation of machine translation}.
\newblock In \emph{Proceedings of the 40th Annual Meeting of the Association
  for Computational Linguistics}, pages 311--318, Philadelphia, Pennsylvania,
  USA. Association for Computational Linguistics.

\bibitem[{Post(2018)}]{post-2018-call}
Matt Post. 2018.
\newblock \href {https://www.aclweb.org/anthology/W18-6319} {A call for clarity
  in reporting {BLEU} scores}.
\newblock In \emph{Proceedings of the Third Conference on Machine Translation:
  Research Papers}, pages 186--191, Belgium, Brussels. Association for
  Computational Linguistics.

\bibitem[{Prabhumoye et~al.(2019)Prabhumoye, Quirk, and
  Galley}]{prabhumoye-etal-2019-towards}
Shrimai Prabhumoye, Chris Quirk, and Michel Galley. 2019.
\newblock \href {https://doi.org/10.18653/v1/N19-1269} {Towards content
  transfer through grounded text generation}.
\newblock In \emph{Proceedings of the 2019 Conference of the North {A}merican
  Chapter of the Association for Computational Linguistics: Human Language
  Technologies, Volume 1 (Long and Short Papers)}, pages 2622--2632,
  Minneapolis, Minnesota. Association for Computational Linguistics.

\bibitem[{Radford et~al.(2019)Radford, Wu, Child, Luan, Amodei, Sutskever
  et~al.}]{radford2019language}
Alec Radford, Jeffrey Wu, Rewon Child, David Luan, Dario Amodei, Ilya
  Sutskever, et~al. 2019.
\newblock Language models are unsupervised multitask learners.
\newblock \emph{OpenAI blog}, 1(8):9.

\bibitem[{Raffel et~al.(2019)Raffel, Shazeer, Roberts, Lee, Narang, Matena,
  Zhou, Li, and Liu}]{raffel2019exploring}
Colin Raffel, Noam Shazeer, Adam Roberts, Katherine Lee, Sharan Narang, Michael
  Matena, Yanqi Zhou, Wei Li, and Peter~J Liu. 2019.
\newblock Exploring the limits of transfer learning with a unified text-to-text
  transformer.
\newblock \emph{arXiv preprint arXiv:1910.10683}.

\bibitem[{Sanh et~al.(2022)Sanh, Webson, Raffel, Bach, Sutawika, Alyafeai,
  Chaffin, Stiegler, Raja, Dey, Bari, Xu, Thakker, Sharma, Szczechla, Kim,
  Chhablani, Nayak, Datta, Chang, Jiang, Wang, Manica, Shen, Yong, Pandey,
  Bawden, Wang, Neeraj, Rozen, Sharma, Santilli, Fevry, Fries, Teehan, Scao,
  Biderman, Gao, Wolf, and Rush}]{sanh2022multitask}
Victor Sanh, Albert Webson, Colin Raffel, Stephen Bach, Lintang Sutawika, Zaid
  Alyafeai, Antoine Chaffin, Arnaud Stiegler, Arun Raja, Manan Dey, M~Saiful
  Bari, Canwen Xu, Urmish Thakker, Shanya~Sharma Sharma, Eliza Szczechla,
  Taewoon Kim, Gunjan Chhablani, Nihal Nayak, Debajyoti Datta, Jonathan Chang,
  Mike Tian-Jian Jiang, Han Wang, Matteo Manica, Sheng Shen, Zheng~Xin Yong,
  Harshit Pandey, Rachel Bawden, Thomas Wang, Trishala Neeraj, Jos Rozen,
  Abheesht Sharma, Andrea Santilli, Thibault Fevry, Jason~Alan Fries, Ryan
  Teehan, Teven~Le Scao, Stella Biderman, Leo Gao, Thomas Wolf, and Alexander~M
  Rush. 2022.
\newblock \href {https://openreview.net/forum?id=9Vrb9D0WI4} {Multitask
  prompted training enables zero-shot task generalization}.
\newblock In \emph{International Conference on Learning Representations}.

\bibitem[{Schick and Sch{\"u}tze(2021)}]{schick-schutze-2021-just}
Timo Schick and Hinrich Sch{\"u}tze. 2021.
\newblock \href {https://doi.org/10.18653/v1/2021.naacl-main.185} {It{'}s not
  just size that matters: Small language models are also few-shot learners}.
\newblock In \emph{Proceedings of the 2021 Conference of the North American
  Chapter of the Association for Computational Linguistics: Human Language
  Technologies}, pages 2339--2352, Online. Association for Computational
  Linguistics.

\bibitem[{Scialom et~al.(2021)Scialom, Dray, Lamprier, Piwowarski, Staiano,
  Wang, and Gallinari}]{scialom-etal-2021-questeval}
Thomas Scialom, Paul-Alexis Dray, Sylvain Lamprier, Benjamin Piwowarski, Jacopo
  Staiano, Alex Wang, and Patrick Gallinari. 2021.
\newblock \href {https://doi.org/10.18653/v1/2021.emnlp-main.529}
  {{Q}uest{E}val: Summarization asks for fact-based evaluation}.
\newblock In \emph{Proceedings of the 2021 Conference on Empirical Methods in
  Natural Language Processing}, pages 6594--6604, Online and Punta Cana,
  Dominican Republic. Association for Computational Linguistics.

\bibitem[{Snover et~al.(2006)Snover, Dorr, Schwartz, Micciulla, and
  Makhoul}]{snover-etal-2006-study}
Matthew Snover, Bonnie Dorr, Rich Schwartz, Linnea Micciulla, and John Makhoul.
  2006.
\newblock \href {https://aclanthology.org/2006.amta-papers.25} {A study of
  translation edit rate with targeted human annotation}.
\newblock In \emph{Proceedings of the 7th Conference of the Association for
  Machine Translation in the Americas: Technical Papers}, pages 223--231,
  Cambridge, Massachusetts, USA. Association for Machine Translation in the
  Americas.

\bibitem[{Wijaya and Yeniterzi(2011)}]{10.1145/2064448.2064475}
Derry~Tanti Wijaya and Reyyan Yeniterzi. 2011.
\newblock \href {https://doi.org/10.1145/2064448.2064475} {Understanding
  semantic change of words over centuries}.
\newblock In \emph{Proceedings of the 2011 International Workshop on DETecting
  and Exploiting Cultural DiversiTy on the Social Web}, DETECT '11, page
  35–40, New York, NY, USA. Association for Computing Machinery.

\bibitem[{Wolf et~al.(2020)Wolf, Debut, Sanh, Chaumond, Delangue, Moi, Cistac,
  Rault, Louf, Funtowicz, Davison, Shleifer, von Platen, Ma, Jernite, Plu, Xu,
  Scao, Gugger, Drame, Lhoest, and Rush}]{wolf-etal-2020-transformers}
Thomas Wolf, Lysandre Debut, Victor Sanh, Julien Chaumond, Clement Delangue,
  Anthony Moi, Pierric Cistac, Tim Rault, Rémi Louf, Morgan Funtowicz, Joe
  Davison, Sam Shleifer, Patrick von Platen, Clara Ma, Yacine Jernite, Julien
  Plu, Canwen Xu, Teven~Le Scao, Sylvain Gugger, Mariama Drame, Quentin Lhoest,
  and Alexander~M. Rush. 2020.
\newblock \href {https://www.aclweb.org/anthology/2020.emnlp-demos.6}
  {Transformers: State-of-the-art natural language processing}.
\newblock In \emph{Proceedings of the 2020 Conference on Empirical Methods in
  Natural Language Processing: System Demonstrations}, pages 38--45, Online.
  Association for Computational Linguistics.

\bibitem[{Zhang et~al.(2020)Zhang, Zhao, Saleh, and Liu}]{zhang2020pegasus}
Jingqing Zhang, Yao Zhao, Mohammad Saleh, and Peter Liu. 2020.
\newblock Pegasus: Pre-training with extracted gap-sentences for abstractive
  summarization.
\newblock In \emph{International Conference on Machine Learning}, pages
  11328--11339. PMLR.

\bibitem[{Zhang* et~al.(2020)Zhang*, Kishore*, Wu*, Weinberger, and
  Artzi}]{Zhang2020BERTScore}
Tianyi Zhang*, Varsha Kishore*, Felix Wu*, Kilian~Q. Weinberger, and Yoav
  Artzi. 2020.
\newblock \href {https://openreview.net/forum?id=SkeHuCVFDr} {Bertscore:
  Evaluating text generation with bert}.
\newblock In \emph{International Conference on Learning Representations}.

\end{thebibliography}
\bibliographystyle{acl_natbib}

\appendix

\section{Details of Prompts}
\label{appendix:prompt}

\paragraph{Other Textual Prompts.}
The three designs of textual prompts we try in our pilot study are:

\begin{enumerate}
    \item \textit{Date: [converted timestamp].}
    \item \textit{Today is [converted timestamp].}
    \item \textit{The following text is written on [converted timestamp].}
\end{enumerate}

As shown in Table~\ref{tab:other_textual_prompt_result}, the second design \textit{``Today is [converted timestamp].''} yields the highest average ROUGE score on the development set of XSum.
Note that performances by the three prompt designs do not vary greatly.

\begin{table}[t]
    \centering
    \small
    \begin{tabular}{lccc}
    \toprule
        \textbf{Prompt Design} & \textbf{R-1} & \textbf{R-2} & \textbf{R-L} \\
        \midrule
        1 (\textit{Date: ...}) & 45.70 & \textbf{22.57} & 37.47 \\
        2 (\textit{Today is ...}) & \textbf{45.71} & 22.55 & \textbf{37.49} \\
        3 (\textit{The following ...}) & 45.61 & 22.54 & 37.32 \\
        \bottomrule
    \end{tabular}
    \caption{ROUGE scores on the dev set of XSum by models with different textual prompts on the encoder. Textual prompt \textit{``Today is [converted timestamp].''} achieves the highest average ROUGE score and is used in our main experiments.}
    \label{tab:other_textual_prompt_result}
\end{table}

\section{Details of Datasets}
\label{appendix:dataset}

\subsection{\data}

\paragraph{Data Collection.}

We use the English Wikipedia dump\footnote{\url{https://dumps.wikimedia.org/}} processed on February 1, 2022 and collect revisions before 2021 that have complete infoboxes.
To identify biographies from all articles, we use the article titles and IDs in \textsc{WikiBio}~\cite{lebret-etal-2016-neural}, which are originally from WikiProject for biographies.\footnote{\url{https://en.wikipedia.org/wiki/Wikipedia:WikiProject_Biography}}
We then extract attributes of the infobox and the first paragraph of the article from each remaining revision with mwparserfromhell.\footnote{\url{https://github.com/earwig/mwparserfromhell}}
Revisions that do not contain complete infoboxes are discarded.
We further discard the first five revisions of each article to avoid including revisions with less comprehensive information about the person.
To limit the number of revisions for each article, we pick the latest revision every X days, where X is sampled uniformly from $[270, 450]$ after picking a revision to diversify the timestamps of selected revisions.

For the test split, the number of articles is downsampled to $10\%$ of the original number of articles created in the test time period to achieve a reasonable running time of decoding.

\paragraph{Time Statistics.}

The numbers of revisions made in different years are shown in Table~\ref{tab:wiki_time_stat}.

\begin{table}[t]
    \centering
    \small
    \setlength{\tabcolsep}{3pt}
    \begin{tabular}{ccccc}
    \toprule
        \textbf{Year} & \textbf{Train} & \textbf{Dev} & \textbf{Test-Same Time} & \textbf{Test-Future} \\
        \midrule
        2004 & 56 & 3 & 1 & - \\
        2005 & 2{,}167 & 118 & 10 & - \\
        2006 & 24{,}509 & 1{,}234 & 132 & - \\
        2007 & 76{,}637 & 3{,}794 & 421 & - \\
        2008 & 133{,}467 & 6{,}770 & 802 & - \\
        2009 & 194{,}504 & 9{,}882 & 1{,}209 & - \\
        2010 & 248{,}768 & 12{,}578 & 1{,}685 & - \\
        2011 & 309{,}642 & 15{,}801 & 2{,}266 & - \\
        2012 & 371{,}008 & 18{,}893 & 2{,}927 & - \\
        2013 & 365{,}260 & 18{,}619 & 3{,}039 & - \\
        2014 & 397{,}155 & 20{,}572 & 3{,}531 & - \\
        2015 & 441{,}800 & 22{,}597 & 4{,}286 & - \\
        2016 & 535{,}693 & 27{,}771 & 5{,}462 & - \\
        2017 & 460{,}233 & 23{,}733 & 4{,}602 & - \\
        2018 & 445{,}955 & 22{,}872 & 4{,}169 & - \\
        2019 & - & - & - & 11{,}698 \\
        2020 & - & - & - & 11{,}998 \\
        2021 & - & - & - & 11{,}132 \\
        \bottomrule
    \end{tabular}
    \caption{The numbers of \data revisions made in different years, grouped by different splits.}
    \label{tab:wiki_time_stat}
\end{table}

\paragraph{Copyright Policy.}

We comply with the Wikipedia copyright policy\footnote{\url{https://en.wikipedia.org/wiki/Wikipedia:Copyrights}} to collect the \data. \data will be released under the CC BY-SA 3.0 license.\footnote{\url{https://creativecommons.org/licenses/by-sa/3.0/}}
The usage of \data is limited by the copyright policy of Wikipedia.

\subsection{Content Transfer}

The content transfer dataset~\cite{prabhumoye-etal-2019-towards} extracts sentences with citations to news outlets in Wikipedia as the sentences to be generated. The cited source documents then become the external source documents where the generation should be grounded. The three sentences preceding each sentence to be generated in the original Wikipedia article are taken as the context passage.
As the source documents come from many different news sources, instead of constructing an extraction template for each new source, we query the Wayback Machine~\footnote{\url{https://web.archive.org}} for the date when each source document was first archived to obtain the timestamp.

\paragraph{Time Statistics.}

In Table~\ref{tab:cf_stat}, we report the numbers of source documents in the content transfer dataset published in different years. 

\begin{table}[t]
    \centering
    \small
    \setlength{\tabcolsep}{5pt}
    \begin{tabular}{lccc}
    \toprule
        \textbf{Year} & \textbf{Train} & \textbf{Dev} & \textbf{Test} \\
        \midrule
        unknown & 1{,}719 & 14 & 250 \\
        1996 & 186 & 1 & 19 \\
        1997 & 100 & 1 & 8 \\
        1998 & 34 & 0 & 3 \\
        1999 & 399 & 3 & 29 \\
        2000 & 1{,}012 & 13 & 92 \\
        2001 & 831 & 11 & 99 \\
        2002 & 3{,}561 & 57 & 294 \\
        2003 & 6{,}672 & 150 & 553 \\
        2004 & 4{,}119 & 63 & 443 \\
        2005 & 4{,}112 & 66 & 441 \\
        2006 & 7{,}323 & 89 & 581 \\
        2007 & 12{,}398 & 151 & 937 \\
        2008 & 17{,}920 & 176 & 1{,}562 \\
        2009 & 26{,}090 & 298 & 2{,}499 \\
        2010 & 30{,}505 & 311 & 2{,}612 \\
        2011 & 40{,}035 & 400 & 3{,}256 \\
        2012 & 71{,}152 & 708 & 6{,}219 \\
        2013 & 94{,}216 & 799 & 7{,}274 \\
        2014 & 84{,}661 & 830 & 7{,}193 \\
        2015 & 55{,}248 & 588 & 5{,}229 \\
        2016 & 49{,}927 & 572 & 4{,}455 \\
        2017 & 44{,}243 & 509 & 3{,}489 \\
        2018 & 22{,}200 & 215 & 2{,}316 \\
        2019 & 1{,}337 & 22 & 147 \\
        Total & 580{,}000 & 5{,}000 & 50{,}000 \\
        \bottomrule
    \end{tabular}
    \caption{Numbers of source documents for content transfer published in different years, grouped by different splits.}
    \label{tab:cf_stat}
\end{table}

\paragraph{Copyright Policy.}

The content transfer dataset is publicly available\footnote{\url{https://www.microsoft.com/en-us/research/project/content-transfer}} with the usage limited by the MIT License.\footnote{\url{https://opensource.org/licenses/MIT}}

\subsection{XSum}

We conduct experiments on text summarization with XSum~\cite{narayan-etal-2018-dont}, which contains articles from BBC News. During the construction of the dataset, the first sentence of each article is taken as the summary of the remaining content.
The timestamp of each news article is extracted from its corresponding HTML file.

\paragraph{Time Statistics.}

In Table~\ref{tab:xsum_stat}, we report the numbers of articles in XSum published in different years. 

\begin{table}[t]
    \centering
    \small
    \setlength{\tabcolsep}{5pt}
    \begin{tabular}{lccc}
    \toprule
        \textbf{Year} & \textbf{Train} & \textbf{Dev} & \textbf{Test} \\
        \midrule
        2009 & 0 & 0 & 1 \\
        2010 & 1{,}142 & 60 & 62 \\
        2011 & 2{,}820 & 154 & 153 \\
        2012 & 5{,}450 & 304 & 319 \\
        2013 & 7{,}939 & 409 & 420 \\
        2014 & 15{,}409 & 810 & 855 \\
        2015 & 49{,}041 & 2{,}792 & 2{,}736 \\
        2016 & 70{,}922 & 3{,}928 & 3{,}983 \\
        2017 & 51{,}322 & 2{,}875 & 2{,}805 \\
        Total & 204{,}045 & 11{,}332 & 11{,}334 \\
        \bottomrule
    \end{tabular}
    \caption{Numbers of XSum articles published in different years, grouped by different splits.}
    \label{tab:xsum_stat}
\end{table}

\paragraph{Copyright Policy.}

XSum dataset is publicly available\footnote{\url{https://github.com/EdinburghNLP/XSum}} with the usage limited by the MIT License.

\section{Additional Results}
\label{appendix:additional_results}

\paragraph{\data.}

We additionally evaluate the model outputs on \data with PARENT~\cite{dhingra-etal-2019-handling} and report the average number of dates in each model output.
As shown in Table~\ref{tab:wiki_additional_result}, the trend of PARENT scores is similar to other metrics, where the model with linear prompts on the encoder achieves the best result on samples drawn at a future time, while the model with textual prompts on the decoder achieves the best result on samples drawn at the same time period of the training and development sets.

\begin{table}[t]
    \centering
    \small
    \begin{tabular}{ccll}
    \toprule
        & \textbf{Prompt} & \textbf{Parent} & \textbf{\# Date} \\
        \midrule
        \multirowcell{5}{\textbf{\textit{Test}}\\\textbf{\textit{Future}}} & - & 56.30 & 1.33 \\
        \cdashline{2-4}
        & \textsc{Enc:T} & \hlc[positiveblue]{56.40} & 1.33 \\
        & \textsc{Enc:L} & \hlc[positiveblue]{\textbf{56.57}}$^\ast$ & 1.33 \\
        & \textsc{Dec:T} & \hlc[positiveblue]{\textit{56.52}}$^\ast$ & 1.35 \\
        & \textsc{Dec:L} & \hlc[positiveblue]{56.36} & 1.32 \\
        \midrule
        \multirowcell{5}{\textbf{\textit{Test}}\\\textbf{\textit{Same}}\\\textbf{\textit{Time}}} & - & 57.57 & 1.25 \\
        \cdashline{2-4}
        & \textsc{Enc:T} & \hlc[positiveblue]{57.74}$^\ast$ & 1.25 \\
        & \textsc{Enc:L} & \hlc[positiveblue]{\textit{57.82}}$^\ast$ & 1.25 \\
        & \textsc{Dec:T} & \hlc[positiveblue]{\textbf{57.88}}$^\ast$ & 1.26 \\
        & \textsc{Dec:L} & 57.57 & 1.24 \\
        \bottomrule
    \end{tabular}
    \caption{
    Additional results on \data. Textual (T) and linear (L) prompts are used on the encoder (\textsc{Enc}) or decoder (\textsc{Dec}). 
    The best result is in \textbf{boldface} and the second best is in \textit{italics}.
    Improvement over the no-prompt baseline is \hlc[positiveblue]{shaded}.
    $^\ast$: significantly better than the baseline with approximate randomization test ($p < 0.001$). 
    }
    \label{tab:wiki_additional_result}
\end{table}

\paragraph{Content Transfer and XSum.}

We report BLEU-4 on content transfer and ROUGE-L on XSum in Table~\ref{tab:cf_additional_result}.
Textual prompts yield the best performance on the two datasets.

\begin{table}[t]
    \centering
    \small
    \begin{tabular}{cllll}
    \toprule
        & \multicolumn{2}{c}{\textbf{Content Transfer}} & \multicolumn{2}{c}{\textbf{XSum}} \\
        \cmidrule(lr){2-3}\cmidrule(lr){4-5}
        \textbf{Prompt} & \textbf{BLEU-4} & \textbf{\# Date} & \textbf{ROUGE-L} & \textbf{\# Date} \\
        \midrule
        - & 11.05 & 0.530 & 37.04 & 0.275 \\
        \hdashline
        \textsc{Enc:T} & \hlc[positiveblue]{\textit{11.52}}$^\ast$ & 0.617 & \hlc[positiveblue]{\textbf{37.34}}$^\ast$ & 0.289 \\
        \textsc{Enc:L} & \hlc[positiveblue]{11.27}$^\ast$ & 0.548 & \hlc[positiveblue]{37.15} & 0.272 \\
        \textsc{Dec:T} & \hlc[positiveblue]{\textbf{11.66}}$^\ast$ & 0.610 & \hlc[positiveblue]{\textbf{37.34}}$^\ast$ & 0.293 \\
        \textsc{Dec:L} & \hlc[positiveblue]{11.10} & 0.536 & 36.84 & 0.287 \\
        \bottomrule
    \end{tabular}
    \caption{Additional results on the content transfer and XSum datasets.
    }
    \label{tab:cf_additional_result}
\end{table}

\paragraph{Sensitivity Analyses.}

\begin{figure}
    \centering
    \includegraphics[width=0.48\textwidth]{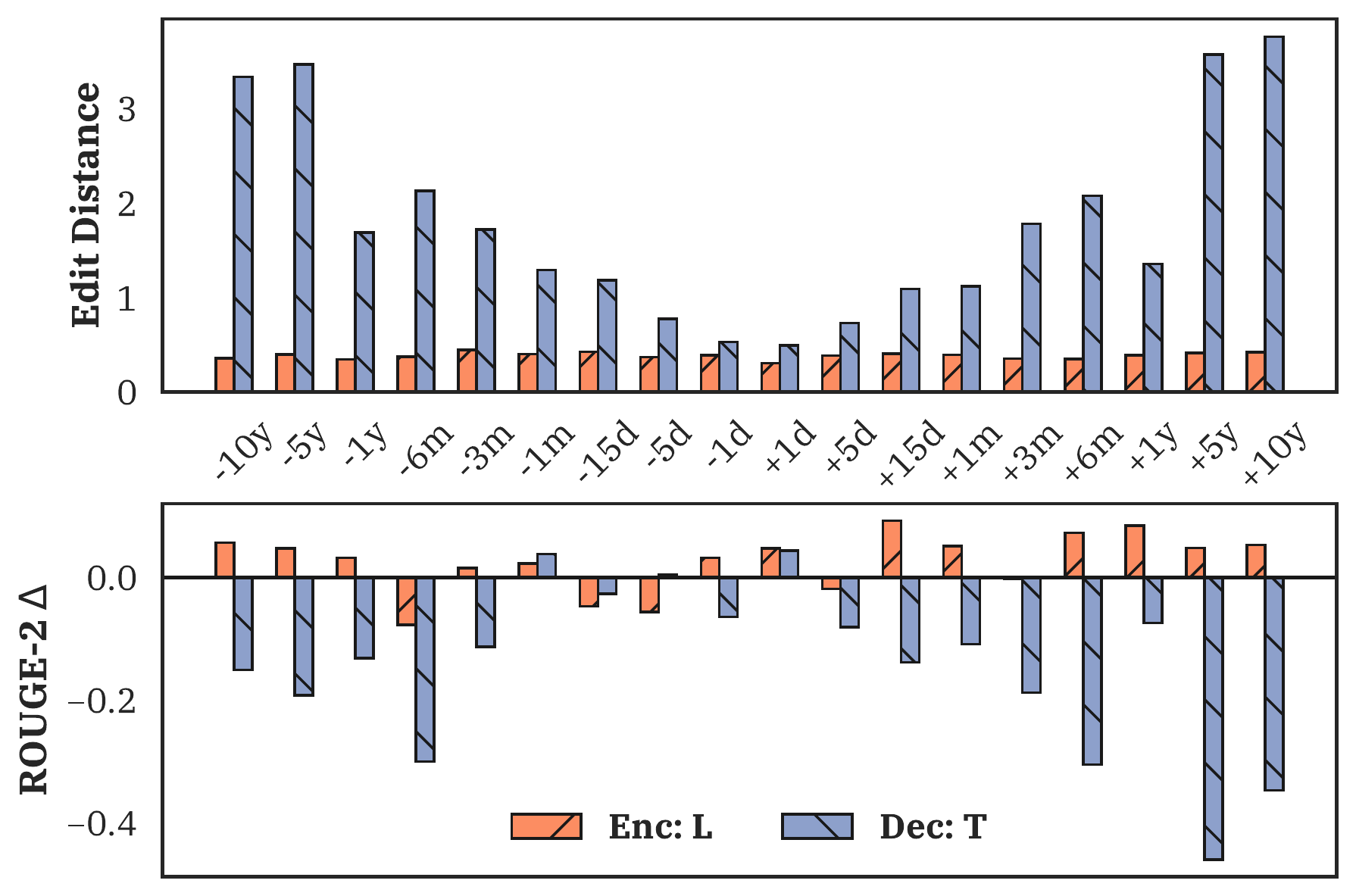}
    \caption{Edit distances and differences of ROUGE-2 between outputs with perturbed dates and original dates on XSum. Similar to the results on \data and content transfer, linear prompts are not sensitive to the given dates.
    }
    \label{fig:intervention_xsum_result}
\end{figure}

We show the results of sensitivity analyses with date perturbation on XSum in Figure~\ref{fig:intervention_xsum_result}.
Linear prompts again do not show sensitivity to the given dates.
Compared to feeding the model with dates that are a year later or earlier, greater drops of ROUGE-2 are observed when feeding the models with dates that are 6 months later or earlier.
This suggests that XSum emphasizes resolving date relations of months and days.

\section{Experiments with T5}
\label{appendix:t5}

We also conduct experiments with the T5 pre-trained model~\cite{raffel2019exploring}.
As \texttt{T5-large} has 370 million more parameters than \texttt{bart.large} that has 400 million parameters, we use \texttt{T5-base}, which has 220 million parameters. 
We do not experiment on XSum, where the performance by T5 is shown to be much lower than BART~\cite{gehrmann-etal-2021-gem}.

Results on \data and content transfer are shown in Tables~\ref{tab:wiki_auto_eval_t5} and \ref{tab:cf_auto_eval_t5}.
While linear prompts still work better on \data samples drawn at a future time and textual prompts work better on text-to-text content transfer,
the improvements by linear prompts are less substantial. We conjecture that T5 is pre-trained with natural language prefixes for multiple tasks and prefers textual prompts.

\begin{table}[t]
    \centering
    \small
    \setlength{\tabcolsep}{4pt}
    \begin{tabular}{clllll}
    \toprule
        & \textbf{Prompt} & \textbf{B-4} ($\uparrow$) & \textbf{MTR} ($\uparrow$) & \textbf{TER} ($\downarrow$) & \textbf{BS} ($\uparrow$) \\
        \midrule
        \multirowcell{3}{\textbf{\textit{Test}}\\\textbf{\textit{Future}}} & - & 25.24 & 50.29 & 70.89 & 42.94 \\
        \cdashline{2-6}
        & \textsc{Enc:T} & 25.19 & 50.27 & 71.03 & 42.86 \\
        & \textsc{Enc:L} & \hlc[positiveblue]{\textbf{25.25}} & \hlc[positiveblue]{\textbf{50.32}} & \hlc[positiveblue]{\textbf{70.79}} & \hlc[positiveblue]{\textbf{42.96}} \\
        \midrule
        \multirowcell{3}{\textbf{\textit{Test}}\\\textbf{\textit{Same}}\\\textbf{\textit{Time}}} & - & 24.33 & 50.00 & 72.30 & 42.54 \\
        \cdashline{2-6}
        & \textsc{Enc:T} & \hlc[positiveblue]{\textbf{24.38}} & \hlc[positiveblue]{\textbf{50.06}} & \hlc[positiveblue]{\textbf{72.19}} & \hlc[positiveblue]{\textbf{42.58}} \\
        & \textsc{Enc:L} & \hlc[positiveblue]{24.35} & \hlc[positiveblue]{50.01} & \hlc[positiveblue]{72.21} & \hlc[positiveblue]{42.57}$^\ast$ \\
        \bottomrule
    \end{tabular}
    \caption{
    Results on \data with T5 as the base model. Textual (T) and linear (L) prompts are used on encoder (\textsc{Enc}) or decoder (\textsc{Dec}). 
    B: BLEU; MTR: METEOR; BS: BERTScore. The best result per metric is in \textbf{boldface}.
    Improvement over the no-prompt baseline is \hlc[positiveblue]{shaded}.
    }
    \label{tab:wiki_auto_eval_t5}
\end{table}

\begin{table}[t]
    \centering
    \small
    \begin{tabular}{cllll}
    \toprule
        \textbf{Prompt} & \textbf{R-L} & \textbf{B-4} & \textbf{MTR} & \textbf{BS} \\
        \midrule
        - & 22.71 & 7.65 & 22.31 & 24.04 \\
        \hdashline
        \textsc{Enc:T} & \hlc[positiveblue]{\textbf{23.17}}$^\ast$ & \hlc[positiveblue]{\textbf{8.00}}$^\ast$ & \hlc[positiveblue]{\textbf{23.11}}$^\ast$ & \hlc[positiveblue]{\textbf{24.73}}$^\ast$  \\
        \textsc{Enc:L} & 22.68 & 7.61 & 22.30 & 23.96 \\
        \bottomrule
    \end{tabular}
    \caption{Results on the content transfer dataset with T5 as the base model. R: ROUGE.
    $^\ast$: significantly better than the baseline with approximate randomization test ($p < 0.001$). 
    }
    \label{tab:cf_auto_eval_t5}
\end{table}
\section{Details of Human Evaluation}

\begin{figure*}[t]
    \centering
    \fontsize{10}{12}\selectfont
    \begin{tabular}{|p{0.92\textwidth}|}
    \hline
        \textbf{Annotation Instruction} \\
        \hline
        The annotation task consists of \textbf{80} groups of paragraphs produced by \textbf{two} systems that briefly describe the career of a person. In addition to the paragraphs, each group includes a infobox listing out important information about the person. You will also find a reference paragraph and a baseline paragraph for each group. \\
        Please read each system-produced paragraph and compare it with the baseline paragraph on two aspects: \textbf{informativeness} and \textbf{factuality}. For each aspect, if the system-produced paragraph is better, please label ``win''; if the system-produced paragraph is worse, please label ``lose''; if the two paragraphs are similar, please label ``tie''. \\
        When you label ``win'' or ``lose'', if the better or worse aspect is due to date mentions, please label ``win (date)'' and ``lose (date)'' correspondingly. \\
        The explanation of the two aspects is shown below along with an example. \\
        \hline
        \textbf{Example} \\
        \hline
        \smallskip
        \includegraphics[width=0.48\textwidth]{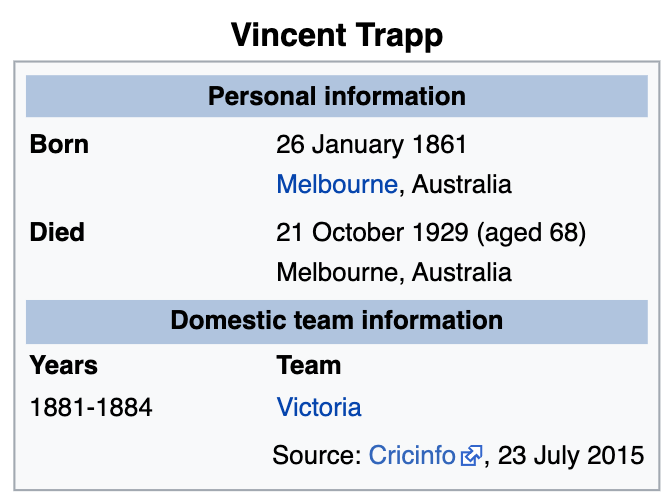} \\
        
        \textbf{Baseline:} Vincent Trapp (26 January 1861) was an Australian cricketer. He played two first-class cricket matches for Victoria between 1881 and 1884. \\
        
        \textbf{System1:} Vincent Trapp (26 January 1861 – 21 October 1929) was an Australian cricketer. He played two first-class cricket matches for Victoria between 1881 and 1884. \\
        
        \textbf{System2:} Vincent Trapp (26 January 1861 – 21 October 1929) was an Australian cricketer. He played for Victoria between 1881 and 1884. \\
        
        \textbf{System3:} Vincent Trapp (26 January 1861) was an Australian cricketer. He played two first-class cricket matches for Victoria between 1881 and 1884, according to Cricinfo. \\

        \smallskip
        \textbf{Informativeness:} Whether the paragraph synthesizes salient information about the person. \\
        In this example, both system 1 and 2 are better than the baseline as they mention the death date of the person, which is an important information, while system 3 that additionally talks about the source of the information ties with the baseline. Moreover, system 1 and system 2 should be labeled with ``win (date)'' as the information is related to a date. \\
        
        \smallskip
        \textbf{Factuality:} Whether the content of the paragraph is factually correct. \\
        In this example, both system 1 and 3 tie with the baseline. System 2 is better than the baseline as it avoids mentioning ``two first-class cricket matches'' which is an incorrect information. System 2 should only be labeled with ``win''. \\
        
        \hline
    \end{tabular}
    \caption{Guideline for human evaluation on \textsc{WikiRevision}.}
    \label{fig:human_eval_guideline}
\end{figure*}

Figures~\ref{fig:human_eval_guideline} and \ref{fig:human_eval_cf_guideline} include the instructions provided to annotators for our human evaluation.
All annotators are college students based in the U.S.
The purpose of the annotation study and the usage of collected data are explained to the annotators before the annotation begins.
We compensate each annotator with \$15 per hour.

\section{Details of Implementation}

For experiments with BART~\cite{lewis-etal-2020-bart}, we use \texttt{bart.large}.\footnote{\url{https://github.com/pytorch/fairseq/tree/main/examples/bart}}
For experiments with T5~\cite{raffel2019exploring}, we use \texttt{T5-base}.\footnote{\url{https://huggingface.co/t5-base}}
Fairseq~\cite{ott2019fairseq}\footnote{\url{https://github.com/pytorch/fairseq/tree/2380a6e4}} is used for model training and decoding with BART. 
HuggingFace Transformer~\cite{wolf-etal-2020-transformers} is used for decoding with T5.
Experiments are conducted with NVIDIA A6000 GPU with 48GB memory.

\paragraph{Training Settings.}

For training on all datasets with BART, we first follow the hyperparameter setting provided by the original BART training script for XSum\footnote{\url{https://github.com/pytorch/fairseq/blob/main/examples/bart/README.summarization.md}} except that we set the total number of update steps to $30{,}000$ for \data and $35{,}000$ for the content transfer dataset.
In addition, we adjust the accumulated batch size for training on \data to have $65{,}536$ tokens in each batch.
We then tune the learning rates on \data and the content transfer dataset by searching through $1\times 10^{-5}$, $3\times 10^{-5}$, and $5\times 10^{-5}$ with the model without prompts. Based on the BLEU-4 scores on the development sets, we choose $5\times 10^{-5}$ for \data and $3\times 10^{-5}$ for the content transfer dataset.
Each model is trained for one run with one random seed due to the high computational cost of fine-tuning large models.
For experiments with T5, we follow the default parameters suggested by HuggingFace.

\paragraph{Decoding Settings.}

We use beam search with beam sizes of 4, 4 and 6 for decoding on \data, content transfer, and XSum. The maximum decoding lengths are set to 100, 100, and 60 for \data, content transfer, and XSum.

\paragraph{Running Time.}

When using 4 GPUs, training on \data, content transfer, and XSum takes 11, 7, and 2 hours. Meanwhile, decoding on \data, content transfer, and XSum respectively takes 2 hours, 1 hour, and 15 minutes with 1 GPU.

\paragraph{Evaluation.}

We use sacreBLEU~\cite{post-2018-call}\footnote{\url{https://github.com/mjpost/sacrebleu}} for calculating the BLEU and TER scores. To obtain the METEOR~\cite{lavie-agarwal-2007-meteor} score, we use NLTK~\cite{bird2009natural}.
The official BERTScore~\cite{Zhang2020BERTScore}\footnote{\url{https://github.com/Tiiiger/bert_score}}, QuestEval~\cite{scialom-etal-2021-questeval}\footnote{\url{https://github.com/ThomasScialom/QuestEval}}, and PARENT~\cite{dhingra-etal-2019-handling}\footnote{\url{https://github.com/google-research/language/tree/master/language/table_text_eval}} libraries are used.
For ROUGE scores~\cite{lin-2004-rouge}, we use the Python implementation by Google.\footnote{\url{https://github.com/google-research/google-research/tree/master/rouge}}

\section{Example Outputs}
\label{appendix:example_output}

In Figures~\ref{fig:example_data2text}, \ref{fig:example_cf}, and \ref{fig:example_xsum}, we show example outputs by different systems on \data, content transfer, and XSum.

\begin{figure}[t]
    \centering
    \small
    \setlength{\tabcolsep}{3pt}
    \begin{tabular}{p{0.48\textwidth}}
    \toprule
        \textbf{Date:} 2020-12-07 
        \textbf{Infobox Attributes:} name[Jan Hellström] birth\_date[February 21, 1960] birthplace[Söderköping , Sweden] position[Striker] national\_years[1986–1989] \\
        \textbf{Reference:} Jan Hellström (born February 21, 1960 in Söderköping, Östergötland) is a former Swedish footballer, who played as a striker. \\
        \textbf{BART}: Jan Hellström (born 21 February 1960 in Söderköping) is a former Swedish footballer. \\
        \textbf{Textual Prompt}: Jan Hellström (born 21 February 1960 in Söderköping) is a former Swedish footballer. \\
        \textbf{Linear Prompt}: Jan Hellström (born 21 February 1960 in Söderköping) is a Swedish former footballer who played as a \hlc[correctgreen!50]{striker}. \\
        \midrule
        \textbf{Date:} 2021-05-30 \textbf{Infobox Attributes:} name[George Daniel Brown] birth\_date[1836] death\_date[1902] birth\_place[London , England] masters[DNP] open[	5th: 1860 , 1863] \\
        \textbf{Reference:} George Daniel Brown (1836–1902) was an English professional golfer. He also specialized as a ball and club maker. Brown finished fifth in the 1860 Open Championship and repeated the feat in the 1863 Open Championship. In total, Brown had three top-10 finishes in The Open Championship. \\
        \textbf{BART}: George Daniel Brown (1836–1902) was an English professional golfer who played in the late 19th century. Brown had \hlc[incorrectred!40]{two} top-10 finishes in the Open Championship. His best performance came in the 1860 Open Championship, when he finished in fifth place. \\
        \textbf{Textual Prompt}: George Daniel Brown (1836–1902) was an English professional golfer. Brown placed fifth in the 1860 and 1863 Open Championships. \\
        \textbf{Linear Prompt}: George Daniel Brown (1836–1902) was an English professional golfer. Brown placed fifth in the 1860 and 1863 Open Championships. \\
        \midrule
        \textbf{Date:} 2021-02-08 \textbf{Infobox Attributes:} name[Robert A. King] coach\_years1[1890] coach\_team1[Wabash] overall\_record[0–3] \\
        \textbf{Reference:} Robert A. King was an American football coach. He was the fifth head football coach at Wabash College in Crawfordsville, Indiana, serving for one season, in 1890, and compiling a record of 0–3. \\
        \textbf{BART}: Robert A. King was the third head college football coach for the Wabash College Little Giants located in Crawfordsville, Indiana and he held that position for the 1890 season. His career coaching record at WABash was 0 wins, 3 losses, and 0 ties. \hlc[incorrectred!40]{As of the conclusion of the 2007 season, this ranks him 18th at Wabat} \\
        \textbf{Textual Prompt}: Robert A. King was the third head college football coach for the Wabash College Little Giants located in Crawfordsville, Indiana and he held that position for the 1890 season. His career coaching record at WABash was 0 wins, 3 losses, and 0 ties. \hlc[incorrectred!40]{As of the conclusion of the 2007 season, this ranks him 18th at Wabat} \\
        \textbf{Linear Prompt}: Robert A. King was an American football coach. He served as the head football coach at Wabash College for one season in 1890, compiling a record of 0–3. \\
        \bottomrule
    \end{tabular}
    \caption{Example system outputs on \data. Textual prompts are on the decoder and linear prompts are on the encoder.}
    \label{fig:example_data2text}
\end{figure}

\begin{figure*}[t]
    \centering
    \small
    \setlength{\tabcolsep}{3pt}
    \begin{tabular}{p{0.88\textwidth}}
    \toprule
        \textbf{Context:} The second season of ''Faking It'', an American single-camera romantic comedy, premiered on September 23, 2014, and concluded on November 2, 2015, on the MTV network. \\
        \textbf{Date:} \hlc[fullorange!60]{2014-06-17} 
        \textbf{Source Document:} A day before its season finale, the comedy from showrunner Carter Covington has earned a sophomore run of 10 episodes. MTV is going to be Faking It for another year. The youth-skewing cable network has renewed the comedy starring Katie Stevens and Rita Volk, The Hollywood Reporter has confirmed. MTV will unspool 10 new episodes -- up two from season one -- in 2015. From showrunner Carter Covington, the half-hour comedy series was the first new scripted entry picked up to series under new network topper Susanne Daniels. "Faking It has proved to be the perfect companion show to Awkward, retaining nearly 90 percent of its lead in each week,” Daniels said in a release announcing the news Monday. “We're excited about Carter Covington's delicious plans for season two.” The comedy, which centers on two best friends who are mistakenly outed as lesbians and catapult to instant popularity, opened in April to 1.17 million total viewers. Through its first seven episodes, the comedy has averaged 948,000 total viewers. MTV says the show is the highest-rated new series launch this year with a 1.5 rating among viewers 12-34 and 1.4 million viewers each week when factoring in three days of delayed viewing. The season finale airs Tuesday. For MTV, Faking It comes as Daniels is looking to double the network's roster of original scripted series. In addition to veterans Awkward and Teen Wolf, MTV will also launch comedy Happyland and dramas Finding Carter and Eye Candy. On the pilot side, MTV is readying its adaptation of Screamand has buzzy book adaptation Shannarain development. \\
        \textbf{Reference}: On June, 2014, the series was renewed for a second season of 10 episodes, which was later extended to 20 episodes. \\
        \textbf{BART}: The series was renewed for a second season of 10 episodes on \hlc[incorrectred!40]{September 23, 2014}. \\
        \textbf{Textual Prompt}: On \hlc[correctgreen!50]{June, 2014}, MTV renewed ''Faking It'' for a second season of 10 episodes. \\
        \textbf{Linear Prompt}: The series was renewed for a second season of 10 episodes on \hlc[incorrectred!40]{September 23, 2014}. \\
        \midrule
        \textbf{Context:} Furthermore, the magazine said of the phrase, "So when your square friend uses it, take a little bit of pleasure in knowing they're referencing a stoner comedy – or a drag reality show referencing a stoner comedy – even if they have no idea." In 2014, VH1 began airing a television show called ''Bye Felicia'', and pop singer Jordin Sparks released a mixtape titled ''\#ByeFelicia''. According to Google Trends, the phrase reached its highest usage in mid-2015. \\
        \textbf{Date:} \hlc[fullorange!60]{2017-12-15}
        \textbf{Source Document:} Outgoing White House official Omarosa Manigault Newman says Roberts' 'Bye, Felicia' dig was "petty" and a "black woman civil war." A link has been sent to your friend's email address. A link has been posted to your Facebook feed. To find out more about Facebook commenting please read the Conversation Guidelines and FAQs Following an appearance on 'Good Morning America' about her resignation from the White House, anchor Robin Roberts muttered the phrase about Omarosa Manigault Newmans. USA TODAY Omarosa Manigault Newman clapped back at Robin Roberts after the Good Morning America host took a dig at her on Thursday's show, telling Inside Edition that her comments were "petty" and akin to "a black woman civil war." The White House public liaison and assistant to the president, who is leaving her post on Jan. 20, had told GMA's Michael Strahan, "When I can tell my story — and it is a profound story — I know the world will want to hear." Later, a skeptical Roberts commented, "She said she has a story to tell? I'm sure she'll be selling that story." Then she invoked Friday's classic two-word dismissal for persons unlikely to be missed by anyone: "Bye, Felicia." Do you know how awful you have to be to annoy Robin Roberts? pic.twitter.com/h0gxAhRFRD For anyone unfamiliar with the expression, we'll let its inventor, Ice Cube, explain: "It's the phrase to get ANYBODY out of your face that's saying something stupid." \\
        \textbf{Reference}: On December 14, 2017, Robin Roberts, a host of ABC TV’s ''Good Morning America'', used the phrase to conclude a segment about Omarosa Manigault Newman's departure from the Presidency of Trump administration staff. \\
        \textbf{BART}:In \hlc[incorrectred!40]{January 2017}, Omarosa Manigault Newman responded to Robin Roberts' use of the phrase on ''Good Morning America'', calling it a "black woman civil war". \\
        \textbf{Textual Prompt}: In \hlc[correctgreen!50]{December 2017}, ''Good Morning America'' host Robin Roberts used the phrase during an interview with Omarosa Manigault Newman about her resignation from the White House. \\
        \textbf{Linear Prompt}: In \hlc[incorrectred!40]{January 2017}, ''Good Morning America'' host Robin Roberts used the phrase to refer to Omarosa Manigault Newman, who had just resigned from her position in the White House. \\
        \bottomrule
    \end{tabular}
    \caption{Example system outputs on the content transfer dataset. Textual prompts are on the decoder and linear prompts are on the encoder.
    The publication dates are frequently required in the outputs.
    }
    \label{fig:example_cf}
\end{figure*}

\begin{figure*}[t]
    \centering
    \small
    \setlength{\tabcolsep}{3pt}
    \begin{tabular}{p{0.88\textwidth}}
    \toprule
        \textbf{Date:} \hlc[fullorange!60]{2016-07-04}
        \textbf{XSum Document:} The cloning of the first animal from an adult cell was a remarkable scientific achievement. It promised new treatments for debilitating diseases. But it also raised fears of cloned human beings, designer babies and a dystopian future. \hlc[fullorange!60]{Twenty years on}, neither the hopes nor the fears have been realised. So what is Dolly's legacy? I first saw Dolly in 1997 at the Roslin Institute just outside Edinburgh. She stood apart from the other sheep in the pens at this agricultural research centre. She stood prouder, her fleece seemed like a lion's mane and there was an aura about her. Dolly's creation had echoes of Mary Shelley's classic novel Frankenstein, in which inanimate tissue was brought to life by electricity. Dolly was created from DNA taken from a cell taken from an sheep. The technique involved putting the DNA into an empty eggshell and then zapping it with electricity. This created an embryo. Researchers at Roslin then implanted the embryo into the womb of a sheep which grew into Dolly - an exact genetic copy of the sheep from which the skin cell was taken. It took 277 attempts to clone Dolly and there were many miscarriages on the way. There were also genuine concerns that it would not be long before cloned humans would be walking the Earth - people would try to clone themselves to achieve a kind of immortality or they might try to resurrect a beloved dead relative. The airwaves were filled with conversations about what it meant to be human, whether the clones would be exactly the same as the person from which they were cloned and what kind of world the scientists were tumbling us into. When I met the researchers at Roslin they were acutely aware of public suspicion. And they knew it was important to be clear, open and honest about their work. Dolly's creator, Prof Sir Ian Wilmut, could not be any more different from fictional scientists such as Dr Frankenstein or indeed Dr Moreau, who developed human-like hybrids in a story by HG Wells. By contrast, Prof Wilmut is a quietly-spoken man whose motivation was to improve animal welfare and human health. This bespectacled scientist with his gentle candour was ideally complemented by the enthusiastic and eloquent Dr Simon Best, who was brought in to commercialise the technology to improve human health ... \\
        \textbf{Reference}: The birth of Dolly the sheep seemed one of those moments in scientific research that would change the world forever. \\
        \textbf{BART}:The birth of Dolly the sheep \hlc[incorrectred!40]{in 1997} sent shockwaves through the scientific community. \\
        \textbf{Textual Prompt}: It is \hlc[correctgreen!50]{20 years} since the birth of Dolly the sheep. \\
        \textbf{Linear Prompt}: It is \hlc[correctgreen!50]{20 years} since the birth of Dolly the sheep. \\
        \midrule
        \textbf{Date:} \hlc[fullorange!60]{2016-04-05} 
        \textbf{XSum Document:} Sunderland's Rachael Laws, 25, has been called up to replace the 32-year-old Liverpool keeper, who has a knee injury. England face Belgium in Rotherham on Friday, before travelling to Zenica to play Bosnia on Tuesday. The Lionesses have won both their qualifying games so far. Goalkeepers: Karen Bardsley (Manchester City), Rachael Laws (Sunderland), Carly Telford (Notts County) Defenders: Laura Bassett (Notts County), Lucy Bronze (Manchester City), Gilly Flaherty (Chelsea), Alex Greenwood (Liverpool), Steph Houghton (Manchester City), Alex Scott (Arsenal), Casey Stoney (Arsenal), Demi Stokes (Manchester City), Amy Turner (Notts County) Midfielders: Katie Chapman (Chelsea), Jordan Nobbs (Arsenal), Jo Potter (Birmingham City), Jill Scott (Manchester City), Fara Williams (Arsenal) Forwards: Eniola Aluko (Chelsea), Karen Carney (Chelsea), Gemma Davison (Chelsea), Toni Duggan (Manchester City), Fran Kirby (Chelsea), Ellen White (Notts County). \\
        \textbf{Textual Prompt}: Manchester City goalkeeper Karen Bardsley has been ruled out of England's Euro \hlc[correctgreen!50]{2017} qualifiers against Belgium and Bosnia-Herzegovina. \\
        \textbf{Perturbed Date:} 2012-04-05 \textbf{Textual Prompt}: Manchester City goalkeeper Karen Bardsley has been ruled out of England's Euro \hlc[incorrectred!40]{2012} qualifiers against Belgium and Bosnia-Herzegovina. \\
        \textbf{Perturbed Date:} 2011-04-05 \textbf{Textual Prompt}: Manchester City goalkeeper Karen Bardsley has been ruled out of England's Euro \hlc[incorrectred!40]{2012} qualifiers against Belgium and Bosnia-Herzegovina. \\
        \textbf{Perturbed Date:} 2021-04-05 \textbf{Textual Prompt}: Manchester City goalkeeper Karen Bardsley has been ruled out of England's Euro \hlc[correctgreen!50]{2021} qualifiers against Belgium and Bosnia-Herzegovina. \\
        \textbf{Perturbed Date:} 2020-04-05 \textbf{Textual Prompt}: Manchester City goalkeeper Karen Bardsley has been ruled out of England's Euro \hlc[incorrectred!40]{2020} qualifiers against Belgium and Bosnia-Herzegovina. \\
        \bottomrule
    \end{tabular}
    \caption{Example system outputs on XSum for text summarization. Textual prompts are on the decoder and linear prompts are on the encoder.
    In the first example, Dolly the sheep was actually born on July 5, 1996.
    In the second example, the Women's Euro is held every four years. Therefore, it could only be Euro 2013, 2017, or 2021.
    }
    \label{fig:example_xsum}
\end{figure*}

\begin{figure*}[t]
    \centering
    \fontsize{10}{12}\selectfont
    \begin{tabular}{|p{0.92\textwidth}|}
    \hline
        \textbf{Annotation Instruction} \\
        \hline
        The annotation task consists of \textbf{80} groups of sentences produced by \textbf{two} systems that continue the given context passage using the information in the given source documents. In addition to the sentences, each group includes the context passage and the source document. You will also find a reference sentence and a baseline sentence for each group. \\
        Please read each system-produced sentence and compare it with the baseline sentence on two aspects: \textbf{informativeness} and \textbf{factuality}. For each aspect, if the system-produced sentence is better, please label ``win''; if the system-produced sentence is worse, please label ``lose''; if the two sentences are similar, please label ``tie''. \\
        When you label ``win'' or ``lose'', if the better or worse aspect is due to date mentions, please label ``win (date)'' and ``lose (date)'' correspondingly. \\
        The explanation of the two aspects is shown below along with an example. \\
        \hline
        \textbf{Example} \\
        \hline
        \smallskip
        \textbf{Context Passage:} The Burleigh Waters Library opened in 1991. For decades a local urban myth maintained that sharks were seen as far south in the canal waterways as Burleigh Waters. Alleged sightings and stories were locally spread, but balanced with scepticism. \\
        \textbf{Source Document:} Publication date: 20 February 2003. The Queensland government has warned people not to swim in coastal canal systems after the second fatal shark attack in as many months on the Gold Coast yesterday. An 84-year-old man from Burleigh Waters died after he was attacked by a 2.5 metre bull whaler while swimming in Burleigh Lake just before 6.30am (AEST) ... \\
        
        \smallskip
        
        \textbf{Baseline:} An 84-year-old man from Burleigh Waters was attacked by a 2.5 metre bull whaler while swimming in Burleigh Lake. \\
        
        \textbf{System1:} An 84-year-old man from Burleigh Waters died after he was attacked by a 2.5 metre bull whaler while swimming in Burleigh Lake. \\
        
        \textbf{System2:} In February 2003, an 84-year-old man from Burleigh Waters died after he was attacked by a 2.5 metre bull whaler while swimming in Burleigh Lake. \\
        
        \textbf{System3:} In 2013, an 84-year-old man from Burleigh Waters died after he was attacked by a 2.5 metre bull whaler while swimming in Burleigh Lake. \\

        \smallskip
        \textbf{Informativeness:} Whether the sentence synthesizes salient information of the source document. \\
        In this example, all systems are better than the baseline as they mention the man died after the attack, which is an important information. Moreover, system 2 should be labeled with ``win (date)'' as it also mentions the date of the event. \\
        
        \smallskip
        \textbf{Factuality:} Whether the content of the sentence is factually correct. \\
        In this example, both system 1 and 2 tie with the baseline. System 3 is worse than the baseline as it mentions an incorrect date. System 3 should be labeled with ``lose (date)''. \\
        
        \hline
    \end{tabular}
    \caption{Guideline for human evaluation on the content transfer dataset.}
    \label{fig:human_eval_cf_guideline}
\end{figure*}

\end{document}